\documentclass[11pt]{article}

\usepackage[letterpaper,margin=1in]{geometry}
\usepackage[T1]{fontenc}
\usepackage[utf8]{inputenc}
\usepackage{amsmath}
\usepackage{amssymb}
\usepackage{lmodern}
\usepackage{biolinum}
\usepackage{microtype}
\usepackage[table]{xcolor}
\usepackage{enumitem}
\usepackage{booktabs}
\usepackage{longtable}
\usepackage{array}
\usepackage{tabularx}
\usepackage{multirow}
\usepackage{graphicx}
\usepackage{float}
\usepackage{placeins}
\usepackage{pdflscape}
\usepackage{rotating}
\usepackage{hyperref}
\usepackage{subcaption}
\usepackage{threeparttable} 
\usepackage[sc]{mathpazo}
\usepackage{inconsolata}
\usepackage[most]{tcolorbox}
\usepackage{xspace}
\usepackage{tikz}
\usetikzlibrary{arrows.meta,positioning,calc}

\hypersetup{
  colorlinks=true,
  linkcolor=rufusorange,
  urlcolor=rufusorange,
  citecolor=rufusorange
}

\definecolor{abstractbgcolor}{RGB}{239,129,40}
\renewenvironment{abstract}
  {\begin{tcolorbox}[
      colframe=abstractbgcolor,
      colback=abstractbgcolor!8,
      boxrule=0.01pt,
      arc=2mm,
      enhanced,
      top=10pt,
      bottom=10pt,
      left=15pt,
      right=15pt,
      width=\textwidth
    ]\large\setlength{\baselineskip}{1.1\baselineskip}\setlength{\parindent}{0.0em}\ignorespaces}
  {\end{tcolorbox}}

\setlist[itemize]{leftmargin=1.5em,itemsep=0.2em,topsep=0.2em}
\renewcommand{\arraystretch}{1.18}
\newcommand{\hd}[1]{{\fontsize{7}{7}\selectfont #1}}

\definecolor{guidegray}{RGB}{75,75,75}
\definecolor{fillblue}{RGB}{25,90,180}
\definecolor{ownerteal}{RGB}{0,110,125}
\definecolor{needred}{RGB}{200,30,30}
\definecolor{rufusorange}{RGB}{239,129,40}
\colorlet{bestk}{rufusorange!5}
\definecolor{reportedgray}{gray}{0.5}

\definecolor{sigverify}{RGB}{232,241,248} % hard verifier
\definecolor{sigmixed}{RGB}{247,241,229}  % verifier + judge
\definecolor{sigmodel}{RGB}{250,235,225}  % learned reward model
\definecolor{signone}{RGB}{242,242,242}   % supervised, no reward
\definecolor{edgegray}{RGB}{110,110,110}

\tikzset{
  stg/.style={rounded corners=2.2pt, draw=black!30, line width=0.45pt,
              align=center, inner xsep=2pt, inner ysep=3.5pt,
              font=\scriptsize, minimum height=1.12cm},
  vfy/.style={stg, fill=sigverify},
  mix/.style={stg, fill=sigmixed},
  non/.style={stg, fill=signone},
  fin/.style={stg, fill=sigmodel, draw=rufusorange, line width=1pt},
  fl/.style={-{Stealth[length=4.2pt,width=3.4pt]}, draw=edgegray, line width=0.65pt},
  dat/.style={font=\tiny\itshape, text=black!55, align=center, inner sep=1.2pt},
  key/.style={rounded corners=1.6pt, draw=black!30, line width=0.4pt,
              minimum width=7pt, minimum height=7pt, inner sep=0pt},
}
\usepackage{sectsty}

\newif\ifinternal
\internaltrue

\title{\sffamily\huge
\raisebox{-0.30em}{\includegraphics[
  height=1.40em,
  trim=180 400 180 300,
  clip
]{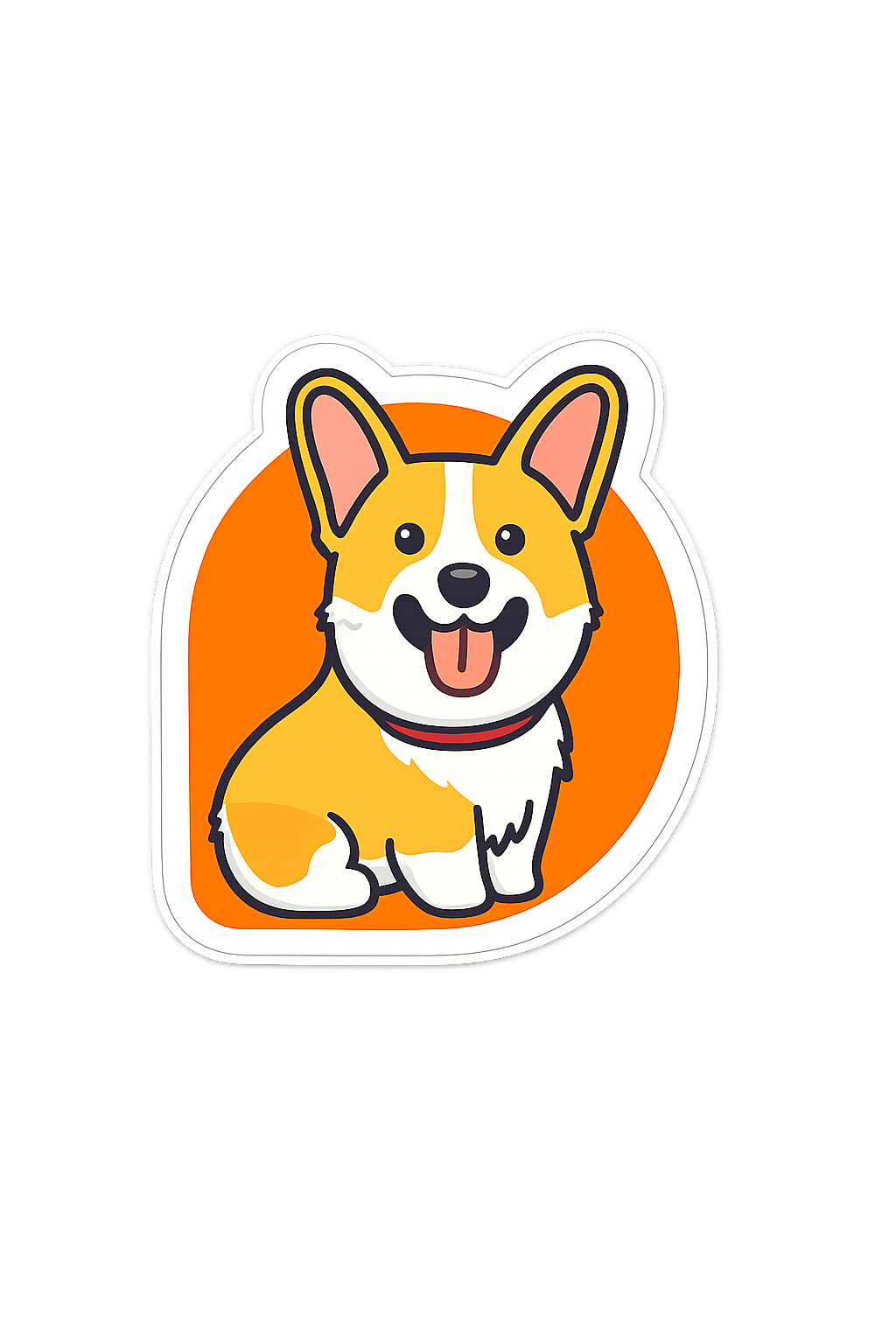}}\hspace{0.00em}%
\textcolor{rufusorange}{Rufus}-Air: An Open LLM Post-Training Recipe}

\author{\sffamily\normalsize Chia-Yuan Chang, Renyuan Cheng, Rui Feng, Xiaotian Han, Yuan He, Hongye Jin,\\\sffamily\normalsize Linwei Li, Shiyang Li, Fenglin Liu, Xin Liu, Priyanka Nigam, Haoyang Wen,\\\sffamily\normalsize Zhenghao Xu, Zhuocheng Xu, Bing Yin, Qingyu Yin, Chao Zhang,\\\sffamily\normalsize Rongzhi Zhang, Zhihan Zhang, Zixuan Zhang$^a$, Zixuan Zhang$^b$, Tuo Zhao\\[0.7em]\sffamily\large Amazon
}
\date{\vspace{-2.2em}}  % no date on the title page; the version date lives with the release

\begin{document}
\maketitle
\footnotetext[1]{\xspace Authors are listed alphabetically by surname; all contributed while at Amazon. Individual contributions are listed in Appendix~\ref{app:contributions}.}
\footnotetext[2]{\xspace The superscripts $^a$ and $^b$ denote two authors with the same romanized name.}

\vspace{-10pt}
\begin{abstract}
Rufus-Air is an open and reproducible post-training recipe on GLM-4.5-Air-Base (106B-A12B), organized as a serial pipeline: SFT $\to$ Reasoning RL $\to$ Coding RL $\to$ Instruction-Following RL $\to$ General Agent $\to$ Coding Agent $\to$ Search Agent $\to$ RLHF. We document the data, reward design, infrastructure, stage order, and stagewise results needed to reproduce the recipe. Stages progress from basic to advanced capabilities and from hard, verifiable rewards to softer judge-based signals. Training builds on open-source components and public data, much of it used as released, without new human annotation or an in-house distillation teacher. Our main findings are that (i) diverse, high-quality SFT establishes a strong capability floor; (ii) difficulty filtering keeps RL prompts within a productive learning range; (iii) reward reliability provides a practical principle for ordering stages; and (iv) infrastructure and engineering choices are part of the recipe, not just an implementation detail. Rufus-Air improves over the official GLM-4.5-Air post-trained release and is competitive with similarly sized open models.
\end{abstract}

% \vspace{10pt}
\begin{center}
\includegraphics[width=0.98\textwidth]{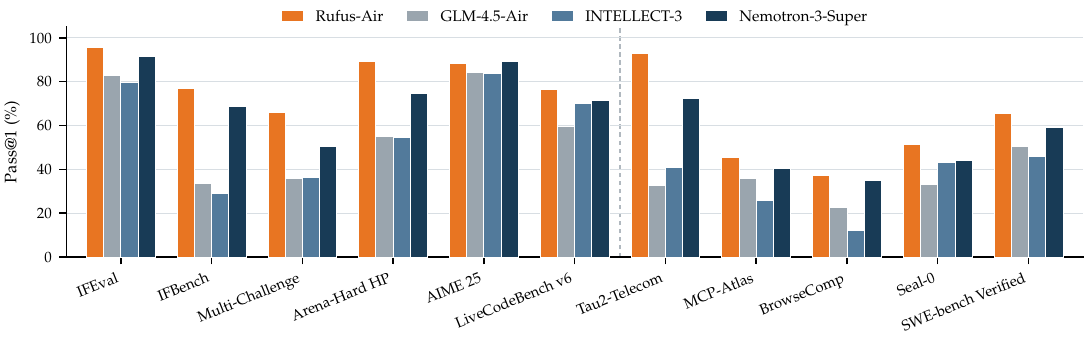}
\end{center}

\newpage
\begingroup\setlength{\parskip}{0.3em}\makeatletter\@starttoc{toc}\makeatother\endgroup

\clearpage

\section{Introduction}\label{sec:introduction}

Open-weight base checkpoints have lowered the barrier to post-training research: many teams can now start from a public model instead of training from scratch. The recipes themselves, however, are still reported thinly, with a few exceptions such as T\"ulu~3~\cite{lambert2024tulu3}. Post-training sections often read more like a system card than like a reproducible recipe, and the details a team would actually need are the least disclosed.

This report reduces that gap with an open, reusable account of one full recipe, \textbf{Rufus-Air}, built on \textbf{GLM-4.5-Air-Base}~\cite{glm45}. The recipe is reproducible for two reasons: it builds entirely on open-source components, and its compute footprint is small enough for teams outside frontier labs (Table~\ref{tab:training_config}). The report sits between a research paper and an engineering experience report. Some of its conclusions come from training experience rather than from full ablations, and we mark them as such (\S\ref{sec:limits}).

At a high level, the recipe is organized around four conclusions, each developed in a later section:
\begin{enumerate}[nosep,leftmargin=1.5em]
  \item \textbf{SFT is a capability-building stage, not a warm-up:} a strong SFT checkpoint establishes broad capabilities that subsequent RL stages refine rather than build from scratch (\S\ref{sec:sft}).
  \item \textbf{Difficulty filtering keeps prompts in a productive band:} we drop prompts the policy already solves and, in most stages, prompts it never solves. Training then focuses on learnable examples~\cite{foster2025lilo}, and the filter acts as an automatic curriculum (\S\ref{sec:reasoning-rlvr}--\S\ref{sec:deep_research}).
  \item \textbf{Reward reliability sets the stage order:} stages with hard, verifiable rewards run first and stages with softer judge- or model-based rewards run later, which limits how long training is exposed to reward hacking. The order follows how easily a reward can be gamed, not its format: IF RL uses a rubric judge yet runs early, because instruction following is close to what the model already does, so the judge has little room to be gamed (\S\ref{sec:pipeline}).
  \item \textbf{Infrastructure is part of the recipe:} the details that made the stages work are rarely reported, among them token-in/token-out rollouts for on-policy multi-turn RL; consistent chat template handling from SFT through the agentic stages; a sandbox service reliable for long agentic runs; and large batches with Rollout Routing Replay~\cite{ma2025r3} for stable RL (\S\ref{sec:train_infra}).
\end{enumerate}
We use benchmarks only to place the recipe among public open models: Rufus-Air leads the official GLM-4.5-Air release on every reported benchmark except Arena-Hard~v2 Creative Writing, and is competitive with open models of similar size such as INTELLECT-3~\cite{intellect2025} and Nemotron-3-Super~\cite{nvidia2026nemotron3super} (Table~\ref{tab:main_comparison}, \S\ref{sec:main_results}). The contribution we care about is the recipe itself: a documented, reusable account that other teams can build on (\S\ref{sec:conclusion}).

\section{Recipe Overview}

This section gives the shape of the whole pipeline before the per-stage detail: how the stages are arranged and why. Figure~\ref{fig:pipeline} summarizes every stage with its targeted capability and reward type.

\medskip\noindent\textbf{Starting checkpoint.} We start from \textbf{GLM-4.5-Air-Base}~\cite{glm45}, a Mixture-of-Experts (MoE) model with 106B total and 12B active parameters, for practical reasons. Its base weights are public. It fits the open-source stack we use: Slime~\cite{slime2025} comes from the group that released GLM-4.5, and SGLang~\cite{zheng2024sglang} supports the GLM family natively. It is large enough that MoE-specific post-training problems show up in realistic form. And with 12B active parameters, the RL stages ran on 8--32 nodes (Table~\ref{tab:training_config}). We use the base model unchanged---architecture, tokenizer, and native \texttt{<think>} chat format---and treat it as fixed. \emph{Rufus-Air} names the checkpoint at the end of the pipeline; an intermediate checkpoint is named after the stage that produced it (e.g., \emph{Rufus-Air SFT}, or \emph{Rufus-Air RL} after instruction-following RL). \emph{GLM-4.5-Air}, the baseline we compare against, is the vendor's own post-trained release, not one of ours.

\subsection{Pipeline Stages and Order}
\label{sec:pipeline}

\begin{figure}[!t]
\centering
\includegraphics[width=0.98\textwidth]{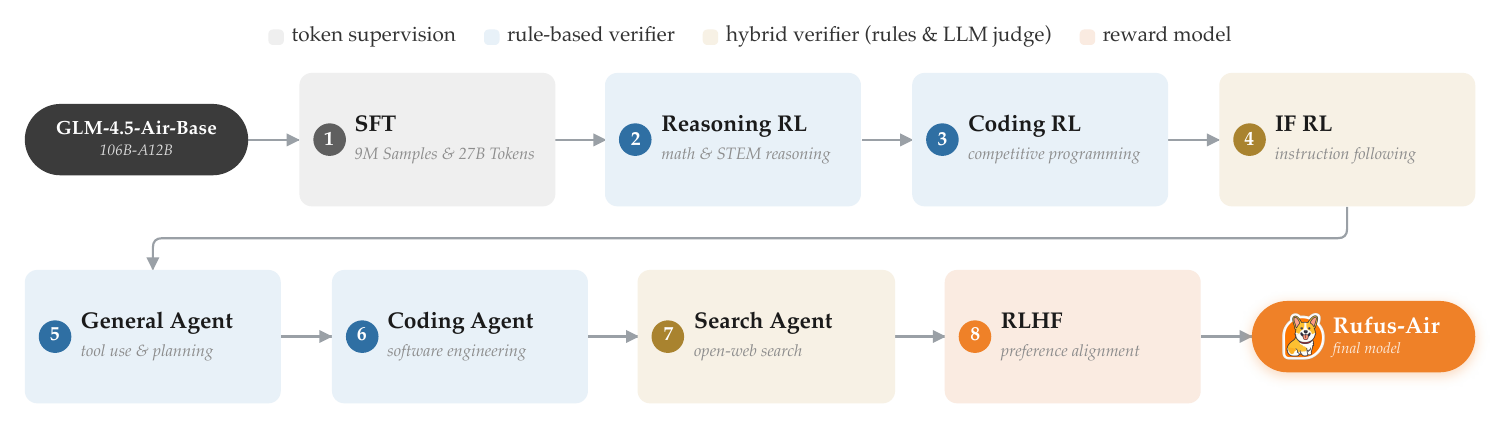}
\vspace{-10pt}
\caption{\textbf{Recipe overview.} The post-training pipeline runs from the \textbf{GLM-4.5-Air-Base} checkpoint through eight stages to the final \textbf{Rufus-Air} model, each stage trained on the checkpoint the previous one produced.
}
\label{fig:pipeline}
\end{figure}
% ChatGPT or CLaude, don't over-complicate this figure, many tech reports keep this clean and precise

\noindent\textbf{Pipeline stages.} The recipe is one serial pipeline: SFT $\to$ Reasoning RL $\to$ Coding RL $\to$ Instruction-Following RL $\to$ General Agent $\to$ Coding Agent $\to$ Search Agent $\to$ RLHF, each stage trained on the checkpoint the previous one produced. SFT builds the base capability and normalizes the output format. Reasoning RL and Coding RL apply reinforcement learning with verifiable rewards (RLVR;~\cite{lambert2024tulu3}) to improve reasoning and coding under low-noise verifiers. IF RL improves exact constraint following and the retention of instructions across turns. The three agentic stages add tool use in their own environments (General Agent, Coding Agent, Search Agent). RLHF~\cite{ouyang2022training} shapes open-ended quality where no hard verifier exists. Figure~\ref{fig:pipeline} shows the stage order and each stage's targeted capability; each stage's data and reward are described in its own subsection.

\medskip\noindent\textbf{Why this order.} Two axes set it. On capability, stages go from basic to advanced, so each builds on what the previous one seeded. On reward type, stages go from hard, verifiable rewards toward softer score-based, judge-based, or environment-mediated signals. Reasoning RL and Coding RL, whose rewards are least vulnerable to hacking, run first; the harder-to-verify stages run later, which shortens the time a gameable reward is under optimization pressure. This order does not protect earlier gains, because every later stage updates the same parameters. We therefore measure each stage against the checkpoint it starts from (Table~\ref{tab:stagewise}); the one exception is SFT, which starts from a base model and is measured against the public GLM-4.5-Air release. The order is not strictly by reward hardness: IF RL uses a rubric-based LLM judge~\cite{zheng2023judging} yet runs before the General Agent and Coding Agent stages, whose rewards are rule-based assertions and execution tests (Figure~\ref{fig:pipeline}). What the order tracks is exposure to reward hacking. Instruction following is close to what the policy already does, so its judge has little room to be gamed, and we saw no serious hacking in this stage; the preference reward of RLHF, a judge on an open-ended objective, is where the risk is real, and that stage runs last.
One local ordering choice, reasoning before instruction following, was inherited from earlier experiments; on GLM-4.5-Air, IF RL gains more from the Reasoning RL checkpoint than from the SFT checkpoint (\S\ref{sec:if-rl}).

% \begin{figure}[!tbp]
% \centering
% \includegraphics[width=0.85\textwidth]{materials/radar-comparison.pdf}
% \vspace{-10pt}
% \caption{\textbf{Rufus-Air against the three same-harness baselines}, pass@1 on every benchmark of Table~\ref{tab:main_comparison} (win rate for the two Arena-Hard~v2 splits). \textbf{(a)}~Non-agentic (instruction following, alignment, reasoning, knowledge); \textbf{(b)}~agentic (general agent, coding agent, search agent). All values are measured under our own harness.}
% \label{fig:radar_comparison}
% \end{figure}

% To ChatGPT or Claude, numbers in Rufus-Air correspond to the "Rufus-Air-v2 (rlhf iter29)" column (the last column of sheet v1) in the GLM Tech Report Results Excel (updated 2026-09-14 from "GLM Tech Report Results (3).xlsx"; previously the 8-2 GLM-Air-Helpfulness-RL-Rank column). GLM-4.5-Air, INTELLECT-3 and Nemotron-3 are all evaluated by ourselves; don't hallucinate or fabricate or replace numbers with public tech reports or something.

\begin{table*}[!t]
\centering
\renewcommand{\arraystretch}{1.48}
\caption{\textbf{Performance comparison.} All values are pass@1 except the two Arena-Hard~v2 rows, which are win rates; \S\ref{sec:eval_setup} describes and cites every benchmark. The \textbf{four leftmost models} we evaluated ourselves under the same harness, and we bold the best of those four in each row; \textcolor{reportedgray}{the other six models} are from other public sources, listed in Appendix~\ref{sec:public_baselines}. A dash marks a benchmark we did not run (first four columns) or that the source did not report (other six).}
\label{tab:main_comparison}
\vspace{-0.9em}
\setlength{\tabcolsep}{2.8pt}
\scriptsize
\begin{tabular}{@{}l *{4}{c} *{6}{>{\color{reportedgray}}c}}
\toprule
\hd{} & \hd{\textbf{\textcolor{rufusorange}{Rufus}-Air}} & \hd{\textbf{GLM-4.5-Air}} & \hd{\textbf{INTELLECT-3}}
& \hd{\textbf{Nemotron-3}} & \hd{Qwen3.5} & \hd{GPT-OSS} & \hd{Ring-flash}
& \hd{Solar-Open} & \hd{Sarvam} & \hd{Mistral} \\
\midrule
Total params & 106B & 106B & 106B & 120B & 122B & 117B & 100B & 102B & 105B & 119B \\
Active params & 12B & 12B & 12B & 12B & 10B & 5B & 6B & 12B & 10B & 7B \\
\midrule
\multicolumn{11}{@{}l}{\emph{Instruction following \& alignment}} \\
\rowcolor{bestk} \shortstack[l]{IFBench\textsubscript{\tiny(prompt strict)}} & \textbf{76.9} & 33.6 & 29.3 & 68.6 & 76.1 & 69.0 & -- & 57.7 & -- & 48.0 \\
\shortstack[l]{IFEval\textsubscript{\tiny(prompt strict)}} & \textbf{95.4} & 83.0 & 79.5 & 91.3 & 93.4 & 88.9 & -- & 88.0 & 84.8 & 84.0 \\
\rowcolor{bestk} Multi-challenge & \textbf{65.8} & 36.0 & 36.5 & 50.7 & 61.5 & 45.3 & -- & 40.5 & -- & -- \\
\shortstack[l]{Arena-Hard v2 (HP) } & \textbf{89.1} & 55.0 & 54.6 & 74.5 & 75.2 & 90.3 & -- & -- & -- & -- \\
\rowcolor{bestk} \shortstack[l]{Arena-Hard v2 (CW) } & 53.0 & 60.3 & 45.4 & \textbf{61.5} & -- & -- & -- & -- & -- & -- \\
\midrule
\multicolumn{11}{@{}l}{\emph{Reasoning \& knowledge}} \\
AIME 25 & 88.3 & 84.2 & 83.8 & \textbf{89.1} & 90.4 & 92.5 & 87.0 & 84.3 & 88.3 & 83.8 \\
\rowcolor{bestk} AIME 26 & \textbf{86.7} & 86.5 & 82.6 & 86.0 & 91.7 & -- & -- & 87.7 & -- & 86.4 \\
LiveCodeBench v6 & \textbf{76.4} & 59.6 & 70.1 & 71.6 & 78.9 & 82.7 & 70.8 & 56.5 & 71.7 & 57.9 \\
\rowcolor{bestk} GPQA & 75.6 & 73.9 & 72.4 & \textbf{75.9} & 86.6 & 80.1 & 75.3 & 66.2 & 78.7 & 71.2 \\
\shortstack[l]{Frontier Sci.(Olympiad)} & 54.6 & 48.1 & 54.3 & \textbf{56.4} & -- & -- & -- & -- & -- & -- \\
\rowcolor{bestk} \shortstack[l]{Frontier Sci.(Research)} & 35.1 & 28.2 & 28.9 & \textbf{35.6} & -- & -- & -- & -- & -- & -- \\
\midrule
\multicolumn{11}{@{}l}{\emph{General Agent}} \\
\rowcolor{bestk} Tau2-Retail & 85.3 & 80.7 & 75.7 & \textbf{86.4} & 62.6 & 76.5 & -- & 59.3 & -- & -- \\
Tau2-Airline & 84.0 & 77.0 & 76.0 & \textbf{86.0} & 66.0 & 56.0 & -- & 52.4 & -- & -- \\
\rowcolor{bestk} Tau2-Telecom & \textbf{93.0} & 32.7 & 40.8 & 72.4 & 95.0 & 57.7 & -- & 55.6 & -- & 41.2 \\
MCP-Atlas & \textbf{45.3} & 35.9 & 26.0 & 40.4 & -- & -- & -- & 34.4 & -- & -- \\
\midrule
\multicolumn{11}{@{}l}{\emph{Search Agent}} \\
\rowcolor{bestk} BrowseComp & \textbf{37.1} & 22.7 & 12.3 & 35.2 & 63.8 & 41.1 & -- & -- & 49.5 & -- \\
Seal-0 & \textbf{51.4} & 33.3 & 43.2 & 44.1 & 44.1 & 45.1 & -- & -- & -- & -- \\
\rowcolor{bestk} HLE-Verified Gold & \textbf{51.1}  & 20.2 & 27.2 & 46.9 & -- & -- & -- & -- & -- & -- \\
\midrule
\multicolumn{11}{@{}l}{\emph{Coding Agent}} \\
\rowcolor{bestk} \shortstack[l]{Terminal-Bench 2.1} & \textbf{42.7} & 24.7 & 25.8 & \textbf{42.7} & 47.6 & 26.2 & -- & -- & -- & 21.0 \\
SWE-bench Verified & \textbf{65.6} & 50.6 & 46.0 & 59.2 & 72.0 & 62.4 & -- & 15.4 & 45.0 & -- \\
\bottomrule
\end{tabular}
\end{table*}

\subsection{Recipe Outcome}
\label{sec:recipe_outcome}

Before the stage-by-stage recipe, Table~\ref{tab:main_comparison} shows where the finished checkpoint lands. We measured three baselines under our own protocol: GLM-4.5-Air and INTELLECT-3~\cite{intellect2025}, which share our base checkpoint, and Nemotron-3-Super~\cite{nvidia2026nemotron3super}, which does not. Against them, Rufus-Air leads on the instruction-following rows and on most of the alignment and agentic rows; the exceptions are Tau2-Airline and, by about a point, Tau2-Retail, where Nemotron-3-Super is ahead, and Arena-Hard~v2 Creative Writing, where both Nemotron-3-Super and GLM-4.5-Air are ahead; on Terminal-Bench~2.1 it ties Nemotron-3-Super (\S\ref{sec:main_results}). On competition mathematics and knowledge, where a single stage, Reasoning RL, does the work, Rufus-Air sits close to the others. In other words, the recipe moves most where it spends the most training signal. The reading caveats behind the table are in \S\ref{sec:eval_setup} and \S\ref{sec:main_results}; the provenance of the six developer-reported columns is in Appendix~\ref{sec:public_baselines}.

% \medskip\noindent\textbf{External models the recipe depends on.} Beyond the base checkpoint, a small set of external models supplies data generation, rewards, and evaluation; reproducers need them to estimate cost and licensing. Correctness-filter teacher for the RLVR stages (Reasoning and Coding): GPT-OSS-120B (\S\ref{sec:rlvr-data}, \S\ref{sec:coding_rlvr}). IF-RL teacher and rubric judge: Qwen3-235B-A22B, with several open models (Qwen3-235B, GPT-OSS-120B, DeepSeek-V3.1, Kimi K2.5) acting as the assistant when multi-turn instructions are created (\S\ref{sec:if-rl}). General-agent correctness-filter teacher: Qwen3-235B-A22B (\S\ref{sec:general_agent}). Search-agent reward judge and learnability filter: Qwen3-32B on dedicated reward-server nodes (\S\ref{sec:deep_research}). RLHF reward model: Skywork-Reward-V2-Qwen3-8B (\S\ref{sec:rm-rl}). Evaluation judge where grading is not deterministic: Claude Sonnet~4.5 via Amazon Bedrock (\S\ref{sec:deep_research}, \S\ref{sec:eval_setup}). Reasoning RL needs no external model for its online difficulty filter, which scores prompts with the current policy itself (\S\ref{sec:reasoning-rlvr}).

\section{Post-Training Stages}
\label{sec:post_training_stages}

This section covers the eight pipeline stages in order (\S\ref{sec:pipeline} and Figure~\ref{fig:pipeline} give their roles at a glance). The first four---SFT, Reasoning RL, Coding RL, and Instruction-Following RL---build general capability; the next three---General Agent, Coding Agent, and Search Agent---add tool use in specialized environments; RLHF closes the pipeline by shaping open-ended quality where no hard verifier exists. Each stage is written to stand on its own---what data entered it, how it was trained, and what came out---and we deliberately keep data and method together rather than splitting them. Throughout, we report the few choices that actually mattered in practice instead of cataloging every training knob. Table~\ref{tab:stagewise} summarizes the outcome: one row per checkpoint, with each stage's change on the benchmarks it targets; the subsections below give the data, training setup, and evaluation behind each row.

\begin{table}[!tbp]
\centering
\scriptsize
\setlength{\tabcolsep}{3pt}
\renewcommand{\arraystretch}{1.15}
\caption{\textbf{Stagewise progression of Rufus-Air}, assembled from the per-stage result tables. One row per checkpoint, in pipeline order: each row carries the benchmarks targeted by the stage that produced it (with the change against the row above) and, in the next stage's target columns, the starting values that stage's table reports for the same checkpoint. Blue cells mark the pairs of numbers each stage's change is read from; \S\ref{sec:main_results} explains how to read the table.}
\label{tab:stagewise}
\vspace{-0.9em}
% Merged variant of tab:stagewise: one row per checkpoint. The "after" row of stage k and the
% "before" row of stage k+1 are the same checkpoint and are merged; their target columns are
% disjoint, so no cell has two sources. Blue = the compared cells; deltas vs the row above.
\scriptsize
\setlength{\tabcolsep}{2.5pt}
% Inline stage deltas: green = up, red = down (relative to the row above).
% Deltas are set in a zero-width box hanging to the right, so the main number stays
% centred under its (bold) column header; tabcolsep is widened to make room for them.
\providecommand{\up}{}\renewcommand{\up}[1]{\makebox[0pt][l]{\,{\fontsize{4.3pt}{5pt}\selectfont\textcolor{green!50!black}{#1}}}}
\providecommand{\dn}{}\renewcommand{\dn}[1]{\makebox[0pt][l]{\,{\fontsize{4.3pt}{5pt}\selectfont\textcolor{red!70!black}{#1}}}}
% The two tinted rows fill up to their rules: the rules touching them are \specialrule with the
% gap on the tinted side set to 0, and a strut in the row restores the height that gap used to add.
\definecolor{cmpblue}{RGB}{228,237,250}  % lighter, slightly bluer grey-blue
% Blue cells are \multicolumn{1}{>{\columncolor{cmpblue}[L][R]}c}{...}: \columncolor takes overhang
% arguments (\cellcolor does not). Interior block edges keep the default \tabcolsep overhang so the
% cells merge; outer block edges use 0.5pt, leaving a 4pt white gap between neighbouring blocks.
% In this copy the deltas are set inside the cell (phantom on the left keeps the number centred).
\providecommand{\dlup}{}\renewcommand{\dlup}[1]{\,{\fontsize{4.3pt}{5pt}\selectfont\textcolor{green!50!black}{#1}}}
\providecommand{\dldn}{}\renewcommand{\dldn}[1]{\,{\fontsize{4.3pt}{5pt}\selectfont\textcolor{red!70!black}{#1}}}
\renewcommand{\arraystretch}{1.5}  % looser rows than tab:stagewise
\resizebox{\textwidth}{!}{%
\begin{tabular}{@{}l ccccccc ccccccc cc@{}}
\toprule
\textbf{Checkpoint} & \textbf{GPQA} & \textbf{AIME25} & \textbf{AIME26} & \textbf{LCBv6} & \textbf{IFEval} & \textbf{IFBench} & \textbf{Multi-ch.} & \textbf{MCP-A} & \textbf{Tau2-Re} & \textbf{TB2.1} & \textbf{SWE-V} & \textbf{BrowseC} & \textbf{Seal-0} & \textbf{HLE-V} & \textbf{AH(HP)} & \textbf{AH(CW)} \\
\specialrule{\lightrulewidth}{\aboverulesep}{0pt}
\rowcolor{black!8} \rule[-1.35ex]{0pt}{4.2ex}GLM-4.5-Air & 73.9 & 84.2 & 86.5 & 59.6 & 83.0 & 33.6 & 36.0 & 35.9 & 80.7 & 24.7 & 50.6 & 22.7 & 33.3 & 20.2 & 55.0 & 60.3 \\
\specialrule{\lightrulewidth}{0pt}{\belowrulesep}
SFT (3799) & \multicolumn{1}{>{\columncolor{cmpblue}[0.5pt][\tabcolsep]}c}{\phantom{\dldn{-5.7}}68.2\dldn{-5.7}} & \multicolumn{1}{>{\columncolor{cmpblue}[\tabcolsep][\tabcolsep]}c}{\phantom{\dlup{+6.6}}90.8\dlup{+6.6}} & \multicolumn{1}{>{\columncolor{cmpblue}[\tabcolsep][0.5pt]}c}{\phantom{\dlup{+3.5}}90.0\dlup{+3.5}} & -- & -- & -- & -- & -- & -- & -- & -- & -- & -- & -- & -- & -- \\
+ Reasoning RL & \multicolumn{1}{>{\columncolor{cmpblue}[0.5pt][\tabcolsep]}c}{\phantom{\dlup{+5.3}}73.5\dlup{+5.3}} & \multicolumn{1}{>{\columncolor{cmpblue}[\tabcolsep][\tabcolsep]}c}{\phantom{\dldn{-2.8}}88.0\dldn{-2.8}} & \multicolumn{1}{>{\columncolor{cmpblue}[\tabcolsep][0.5pt]}c}{\phantom{\dldn{-2.6}}87.4\dldn{-2.6}} & \multicolumn{1}{>{\columncolor{cmpblue}[0.5pt][0.5pt]}c}{68.6} & -- & -- & -- & -- & -- & -- & -- & -- & -- & -- & -- & -- \\
+ Coding RL & -- & -- & -- & \multicolumn{1}{>{\columncolor{cmpblue}[0.5pt][0.5pt]}c}{\phantom{\dlup{+7.3}}75.9\dlup{+7.3}} & \multicolumn{1}{>{\columncolor{cmpblue}[0.5pt][\tabcolsep]}c}{90.5} & \multicolumn{1}{>{\columncolor{cmpblue}[\tabcolsep][\tabcolsep]}c}{63.8} & \multicolumn{1}{>{\columncolor{cmpblue}[\tabcolsep][0.5pt]}c}{31.1} & -- & -- & -- & -- & -- & -- & -- & -- & -- \\
+ IF RL & -- & -- & -- & -- & \multicolumn{1}{>{\columncolor{cmpblue}[0.5pt][\tabcolsep]}c}{\phantom{\dlup{+4.0}}94.5\dlup{+4.0}} & \multicolumn{1}{>{\columncolor{cmpblue}[\tabcolsep][\tabcolsep]}c}{\phantom{\dlup{+14.0}}77.8\dlup{+14.0}} & \multicolumn{1}{>{\columncolor{cmpblue}[\tabcolsep][0.5pt]}c}{\phantom{\dlup{+24.7}}55.8\dlup{+24.7}} & \multicolumn{1}{>{\columncolor{cmpblue}[0.5pt][\tabcolsep]}c}{35.0} & \multicolumn{1}{>{\columncolor{cmpblue}[\tabcolsep][0.5pt]}c}{74.0} & -- & -- & -- & -- & -- & -- & -- \\
+ General Agent & -- & -- & -- & -- & -- & -- & -- & \multicolumn{1}{>{\columncolor{cmpblue}[0.5pt][\tabcolsep]}c}{\phantom{\dlup{+7.8}}42.8\dlup{+7.8}} & \multicolumn{1}{>{\columncolor{cmpblue}[\tabcolsep][0.5pt]}c}{\phantom{\dlup{+9.8}}83.8\dlup{+9.8}} & \multicolumn{1}{>{\columncolor{cmpblue}[0.5pt][\tabcolsep]}c}{38.8} & \multicolumn{1}{>{\columncolor{cmpblue}[\tabcolsep][0.5pt]}c}{65.6} & -- & -- & -- & -- & -- \\
+ Coding Agent & -- & -- & -- & -- & -- & -- & -- & -- & -- & \multicolumn{1}{>{\columncolor{cmpblue}[0.5pt][\tabcolsep]}c}{\phantom{\dlup{+1.4}}40.2\dlup{+1.4}} & \multicolumn{1}{>{\columncolor{cmpblue}[\tabcolsep][0.5pt]}c}{\phantom{\dlup{+2.2}}67.8\dlup{+2.2}} & \multicolumn{1}{>{\columncolor{cmpblue}[0.5pt][\tabcolsep]}c}{34.2} & \multicolumn{1}{>{\columncolor{cmpblue}[\tabcolsep][\tabcolsep]}c}{48.6} & \multicolumn{1}{>{\columncolor{cmpblue}[\tabcolsep][0.5pt]}c}{47.7} & -- & -- \\
+ Search Agent & -- & -- & -- & -- & -- & -- & -- & -- & -- & -- & -- & \multicolumn{1}{>{\columncolor{cmpblue}[0.5pt][\tabcolsep]}c}{\phantom{\dlup{+3.0}}37.2\dlup{+3.0}} & \multicolumn{1}{>{\columncolor{cmpblue}[\tabcolsep][\tabcolsep]}c}{\phantom{\dlup{+5.4}}54.0\dlup{+5.4}} & \multicolumn{1}{>{\columncolor{cmpblue}[\tabcolsep][0.5pt]}c}{\phantom{\dlup{+3.4}}51.1\dlup{+3.4}} & \multicolumn{1}{>{\columncolor{cmpblue}[0.5pt][\tabcolsep]}c}{83.1} & \multicolumn{1}{>{\columncolor{cmpblue}[\tabcolsep][0.5pt]}c}{38.6} \\
+ RLHF & -- & -- & -- & -- & -- & -- & -- & -- & -- & -- & -- & -- & -- & -- & \multicolumn{1}{>{\columncolor{cmpblue}[0.5pt][\tabcolsep]}c}{\phantom{\dlup{+6.0}}89.1\dlup{+6.0}} & \multicolumn{1}{>{\columncolor{cmpblue}[\tabcolsep][0.5pt]}c}{\phantom{\dlup{+14.4}}53.0\dlup{+14.4}} \\
\specialrule{\lightrulewidth}{\aboverulesep}{0pt}
\rowcolor{rufusorange!15} \rule[-1.35ex]{0pt}{4.2ex}Rufus-Air & 75.6 & 88.3 & 86.7 & 76.4 & 95.4 & 76.9 & 65.8 & 45.3 & 85.3 & 42.7 & 65.6 & 37.1 & 51.4 & 51.1 & 89.1 & 53.0 \\
\specialrule{\heavyrulewidth}{0pt}{0pt}
\end{tabular}}
\end{table}

\subsection{Supervised Fine-Tuning}\label{sec:sft}

\textbf{A strong, diverse SFT stage elicits much of the model's capability and sets the floor RL builds on.} SFT is not merely a warm-up. RL works best when the model already has three things: stable reasoning and response formats that verifiers can parse; broad coverage across chat, mathematics, STEM, coding, and tool use; and enough multi-turn and long-context ability that later rewards refine behavior rather than teach it from scratch. We therefore optimize SFT for breadth, format consistency, and long-horizon behavior, aiming for a checkpoint that is already competitive with the public GLM-4.5-Air release. RL can then spend its budget on further improvement rather than on basic formatting and coverage.

\begin{figure}[!tbp]
\centering
\begin{minipage}[t]{0.59\linewidth}
\centering
\vspace{0pt}
\scriptsize
\setlength{\tabcolsep}{3.5pt}
\renewcommand{\arraystretch}{1.15}
\resizebox{\linewidth}{!}{%
\begin{tabular}{@{}lrrrr@{}}
\toprule
\textbf{Category} & \textbf{Samples} & \textbf{\%Sample} & \textbf{Train tokens} & \textbf{\%Token} \\
\midrule
General Agent   & 3{,}568{,}388 & 39.6\% &  3.84B & 14.2\% \\
General Chat    & 1{,}599{,}070 & 17.7\% &  3.30B & 12.2\% \\
STEM            & 1{,}359{,}334 & 15.1\% &  2.37B &  8.8\% \\
Math            & 1{,}146{,}108 & 12.7\% &  7.96B & 29.4\% \\
Code            &   792{,}807 &  8.8\% &  4.25B & 15.7\% \\
Coding Agent    &   549{,}222 &  6.1\% &  5.32B & 19.7\% \\
\midrule
\textbf{Total}  & \textbf{9{,}014{,}929} & \textbf{100\%} & \textbf{27.04B} & \textbf{100\%} \\
\bottomrule
\end{tabular}%
}
\vspace{0.1em}
\captionof{table}{SFT mixture by capability category. Training tokens count only assistant tokens after masking system, user, and tool-observation turns.}
\label{tab:sft_composition}
\end{minipage}\hfill
\begin{minipage}[t]{0.39\linewidth}
\centering
\vspace{3.1pt}
\includegraphics[width=\linewidth]{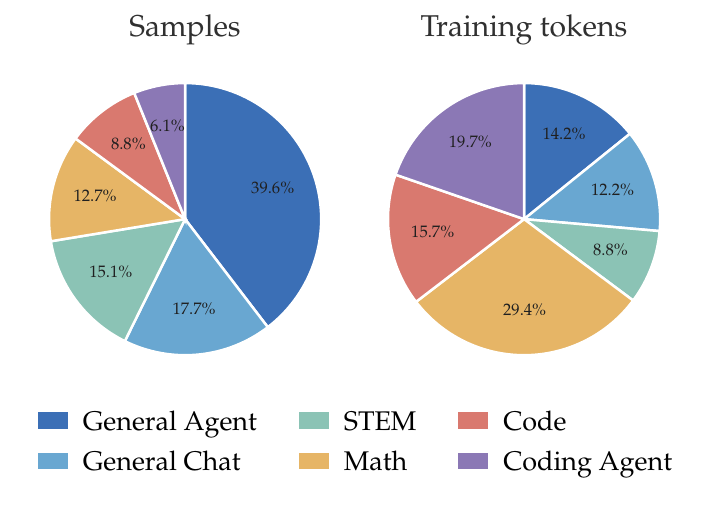}
% \vspace{0.02pt}
\captionof{figure}{SFT composition by sample share (left) and training-token share (right).}
\label{fig:sft_composition_pies}
\end{minipage}
\end{figure}

\paragraph{Data.}
Our SFT data pipeline has three parts: composition, which fixes what goes into the mix; preprocessing, which converts every source into one supervision format; and decontamination, which screens the mix against the benchmarks we report.

\begin{itemize}
  \item \textbf{Data composition.} The final SFT mix contains \textbf{9.01M samples and 44.5B raw tokens}, distributed across \textbf{66.7M conversational turns} (23.5M supervised). After masking system messages, user turns, and tool observations, \textbf{27.0B tokens (60.8\%)} contribute to the training loss. We group the data into six capability categories---\emph{General Agent}, \emph{General Chat}, \emph{STEM}, \emph{Math}, \emph{Code}, and \emph{Coding Agent}---and report both sample and training-token shares in Table~\ref{tab:sft_composition} and Figure~\ref{fig:sft_composition_pies}. The two distributions differ substantially. General Agent is the largest category by sample count but contributes far fewer training tokens. In contrast, Math and Coding Agent together account for 18.8\% of samples but 49.1\% of training tokens, reflecting the greater length of reasoning traces and multi-turn coding trajectories. Reporting both views is therefore important: sample counts describe source coverage, whereas loss-contributing tokens more closely describe the supervision volume seen during optimization. Every prompt and response in the mix comes from one of the 17 public datasets listed in Appendix~\ref{sec:appendix_data}; we use the released records as distributed and do not run a separate Rufus-Air response-regeneration pass or commission new human annotation. Thus, the fraction of the mix regenerated by Rufus-Air is zero. Several source datasets already contain model-generated responses or trajectories: ToolMind~\cite{yang2025toolmind} uses DeepSeek-V2-Chat~\cite{deepseek2024v2}, Mixtral-8x22B-Instruct~\cite{mistral2024mixtral8x22b}, and DeepSeek-V3~\cite{deepseek2024v3}; tool-use-multiturn-reasoning~\cite{interstellarninja_toolusemultiturn} uses DeepSeek-R1~\cite{deepseek2025r1} and QwQ-32B~\cite{qwen2025qwq32b}; ToolMind-Web-QA uses MiroThinker~\cite{miromind2025mirothinker} with Qwen3-235B-A22B-Thinking-2507~\cite{yang2025qwen3}; Toucan-1.5M~\cite{xu2025toucan} uses Qwen3-32B, Kimi-K2~\cite{kimi2025k2}, and GPT-OSS~\cite{openai2025gptoss}; AM-Thinking-v1-Distilled~\cite{tian2025notall} uses AM-Thinking-v1~\cite{ji2025amthinking}; AReaL-tau2-data~\cite{gao2026selfevolving} uses Qwen3-30B-A3B; Superior-Reasoning-SFT-gpt-oss-120b~\cite{yan2026dasd} and OpenResearcher-Dataset~\cite{li2026openresearcher} use GPT-OSS-120B; OpenSeeker-v1-30B-SFT~\cite{du2026openseeker} uses Qwen3-30B-A3B-Thinking-2507; and several aggregated sources include traces from DeepSeek-R1 or QwQ-32B. These are upstream generation dependencies of the public releases, not in-house distillation teachers.
  \item \textbf{Data preprocessing.} Heterogeneous source shards are converted to one supervision format in three steps. (i) \emph{format normalization}: every source is mapped to a role-aware multi-turn schema with a per-message loss mask; source-specific passes repair system prompts, answer suffixes, malformed reasoning tags, and incomplete final turns, and records with no supervised assistant response are dropped. (ii) \emph{interleaved thinking supervision}: agent trajectories alternate between \texttt{<think>} reasoning, tool calls, and observations. We keep each complete thought--action--observation chain as one training target: an assistant turn counts as intermediate only if it issues a tool call that a tool result immediately follows; otherwise it ends the response, and we split there, so each completed response becomes one example. Earlier dialogue stays as context with its assistant rationales stripped and masked; from the last real user request onward, reasoning-bearing assistant steps are supervised and system, user, and tool-observation messages receive zero loss. (iii) \emph{validation and length filtering}: rule-based validators check balanced \texttt{<think>} tags, valid role transitions, and consistent masking around tool interactions. Sequence length is measured after applying the production GLM chat template, so it includes role delimiters, reasoning wrappers, tool schemas, calls, and observations, and is capped at 120K tokens.
  \item \textbf{Data decontamination.} To prevent evaluation data leakage, we screen the SFT mix against reported benchmarks with a word-level 8-gram overlap test: for every benchmark item we sample ${\sim}20$ evenly spaced 8-word phrases and substring-match them against the lowercased concatenation of all conversation turns (system, user, assistant, and tool messages), then drop a sample once $\geq 50\%$ of any single item's phrases hit. We chain the ten passes sequentially --- Terminal-Bench~2.0, HLE~\cite{phan2026benchmark}, HLE-Verified~\cite{zhai2026hle}, Tau2-Bench, SealQA, SWE-bench Verified, Multi-challenge, IFEval, IFBench, and MCP-Atlas --- and remove \textbf{3{,}529 samples in total}: 3{,}321 for Terminal-Bench~2.0 (all from a single synthetic text-to-terminal corpus), 57 for HLE-Verified, 21 for IFEval, and 130 for IFBench; the remaining six benchmarks have zero matches. We inspected abnormally high match counts manually and kept the false positives: e.g., 11{,}874 SWE-bench candidates trace to generic \texttt{pytest} scaffolding of a single task, which we judged benign. Math and knowledge benchmarks (AIME~24 / 25 / 26~\cite{maa_aime}; HMMT February~2025, November~2025, and February~2026; and GPQA-Diamond) are the ones most prone to leakage, so we screen them with two further detectors: (i) an exhaustive 8-gram variant that checks \emph{all} 8-word phrases of every problem (no sampling), and (ii) dense retrieval with Llama-NV-Embed-Reasoning-3B~\cite{nvidia2026nvembedreasoning} embeddings, after which we manually check every training sample with high cosine similarity to any benchmark problem. The exhaustive n-gram screen finds 26 contaminated pairs (25 AIME 24, 1 AIME 26); manual review of the dense candidates confirms a further 42, for \textbf{68 confirmed contaminated pairs (58 unique samples)}, dominated by verbatim or lightly reformatted AIME~24 problems. Together with all borderline cases, this conservative removal drops \textbf{658 unique samples}, bringing the total number of instances removed from the SFT mix to \textbf{4{,}187}. Together these checks make large-scale contamination very unlikely, though, as with any n-gram and embedding filter, they cannot rule out all leakage. The same residual risk applies to every stage's prompt set, not only the SFT mix, because we run this same pipeline over each RL prompt set as we assemble it (\S\ref{sec:rlvr-data}, \S\ref{sec:coding_rlvr}).
\end{itemize}

% We apply the same screen to the RL prompt sets of the later stages, not only to the SFT mix: the chained n-gram passes, the exhaustive math and knowledge variant, and the dense-retrieval check form one pipeline, which we run over each stage's prompt set as we assemble it (\S\ref{sec:rlvr-data}, \S\ref{sec:coding_rlvr}). We report the removal counts here because the SFT mix is the largest and the one where the filter does most of its work; the residual-risk statement above covers every stage's prompt set, not just this one.

\begin{figure}[t]
    \centering
    % New figure 2026-09-16: GLM-4.5-Air-Base_sft-009-2 run (xth's W&B account `ahxt`), rendered by
    % tmp/render_sft_dynamics.py from tmp/wandb_sft009-2_{history,evals}.csv. The previous file,
    % materials/sft-training-loss.pdf (tmp/make_sft_loss_plot.py), was the Phoenix phx-sft-128k-64n
    % run's loss and is kept on disk unchanged.
    \includegraphics[width=0.98\linewidth]{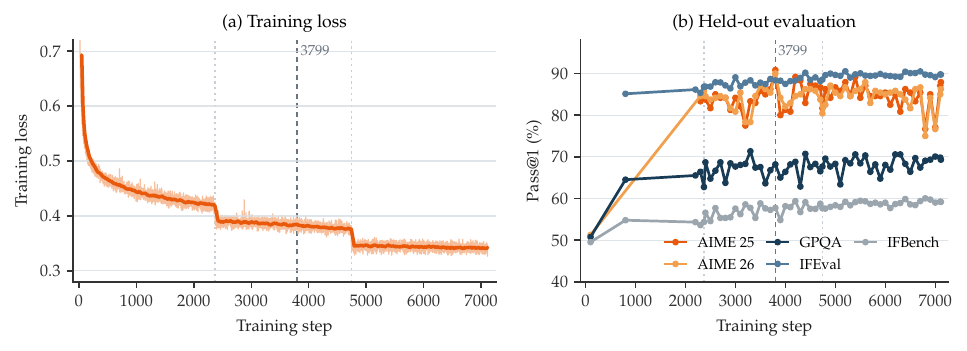}\vspace{-6pt}
    \caption{\textbf{SFT training dynamics} of the shipped SFT run. (a)~Token-level training loss per step (light) and its 50-step moving average (dark). (b)~Pass@1 of the checkpoint saved every 100 steps on AIME~25 and AIME~26 ($n{=}8$, Math-Verify~\cite{mathverify2024} scoring), GPQA Diamond, and IFEval and IFBench (prompt-level strict). The dashed line marks checkpoint~3799, the one the RL pipeline starts from. The loss steps down at each epoch boundary as the data repeats, but the held-out scores flatten within the first epoch, so later checkpoints buy lower loss without better evaluation.}
    \label{fig:sft_training_loss}
\end{figure}

\paragraph{Training setup.}
We start from GLM-4.5-Air-Base and train for three epochs over the full 9.01M-example mixture for SFT. Training uses Slime with the Megatron~\cite{shoeybi2019megatron} backend on 64 8$\times$H200 nodes (512 GPUs total), with a batch size of 4096 sequences. We use AdamW~\cite{loshchilov2019adamw} ($\beta_1=0.9$, $\beta_2=0.95$, $\epsilon=10^{-8}$), weight decay 0.1, and gradient clipping at 1.0. The learning rate warms up linearly for 100 steps to $5\times10^{-5}$ and then follows a cosine schedule toward $5\times10^{-6}$. The peak was set by a preliminary sweep at smaller batch sizes: peaks of $3\mathrm{e}{-4}$ and $5\mathrm{e}{-4}$ were consistently unstable, while runs at $\le 1\mathrm{e}{-4}$ produced monotone loss decreases. The model context length is 128K tokens after packing. Only assistant tokens contribute to the token-level loss. As shown in Figure~\ref{fig:sft_training_loss}, the loss falls from $0.84$ at the first step to a 50-step moving average of $0.42$ at the end of the first epoch (step 2370), then steps down at each epoch boundary to $0.38$ and finally $0.34$ at step 7112, the signature of a model re-seeing repeated data. The held-out scores in panel~(b) do not follow the loss: they flatten within the first epoch, and the remaining epochs move them only within evaluation noise. We therefore carry checkpoint 3799, taken inside that plateau rather than at the final step, into the RL stages.

\begin{table}[!tbp]
\centering
\small
\setlength{\tabcolsep}{7pt}
\renewcommand{\arraystretch}{1.15}
\caption{\textbf{SFT-only checkpoint vs.\ the public GLM-4.5-Air release.} All values are pass@1. We compare SFT checkpoint~3799, the checkpoint the RL pipeline starts from (Table~\ref{tab:reasoning_rl_results}), before any RL, against the public release (RL-included); $\Delta$ is the SFT checkpoint minus the public release. The SFT checkpoint already leads on single-turn instruction following and on both AIME years; it trails only on GPQA, one of the gaps the RL stages then target.}
\label{tab:sft_results}
\vspace{-0.9em}
% Checkpoint 3799 column = W&B values supplied by xth 2026-09-16 (runs of 2026-05-25 and
% 2026-05-29 on _hf_0003799): aime2025 pass_at_1_mathverify 0.9083 (05-29 15:52Z),
% aime2026 0.9000 (05-29 16:28Z), gpqa_diamond_thinking 0.6818 (05-25 18:47Z),
% ifeval_thinking strict pass@1 0.8833 (05-25 16:53Z), ifbench_thinking strict pass@1 0.5775
% (05-25 19:16Z). NOT the 05-31 re-run on disk (AIME25 86.25 / AIME26 88.75 / GPQA 69.07 /
% IFEval 87.96 / IFBench 58.87), and not the 05-25 .bak AIME25 0.7375 (different sampling:
% T 1.0 / top_p 1.0 / top_k -1). Same-condition AIME25 re-runs differ by 4.6 pt (n=8, 30 problems).
% The 6599 column (IFEval 90.60, IFBench 59.80, Multi-challenge 34.07, GPQA 68.40, AIME25 86.50,
% AIME26 85.70, LCB 61.71; xlsx col "(1) GLM-Air-Thinking-sft-6599") was removed 2026-09-16 on
% xth's request, together with the Multi-challenge and LiveCodeBench rows that 3799 was not run on.
\begin{tabular}{@{}lccccc@{}}
\toprule
& \textbf{IFEval} & \textbf{IFBench} & \textbf{GPQA} & \textbf{AIME 25} & \textbf{AIME 26} \\
\midrule
GLM-4.5-Air & 83.00 & 33.60 & 73.90 & 84.20 & 86.50 \\
Rufus-Air SFT (3799)   & 88.33 & 57.75 & 68.18 & 90.83 & 90.00 \\
\addlinespace[2pt]
\midrule
$\Delta$ & $+5.33$ & $+24.15$ & $-5.72$ & $+6.63$ & $+3.50$ \\
\bottomrule
\end{tabular}
\end{table}

\paragraph{What SFT gives the rest of the recipe.}
Checkpoint 3799 is the starting policy of the RL sequence: Reasoning RL initializes from it, and every later stage inherits the result through the intervening checkpoints (\S\ref{sec:pipeline}). It also fixes the message schema and the reasoning and tool-call formats that downstream verifiers and judges parse, and it gives sparse verifier rewards enough initial successes to reinforce (Table~\ref{tab:sft_results}).

\paragraph{Result anchor.}
The SFT-only checkpoint is already a useful standalone model, but its strengths are uneven (Table~\ref{tab:sft_results}). It beats the public GLM-4.5-Air post-trained reference on single-turn instruction following by wide margins, is ahead on both AIME years, and trails on GPQA. Multi-turn instruction following was not measured on this checkpoint, but the checkpoint the IF RL stage starts from sits at 31.1 on Multi-challenge (Table~\ref{tab:if-rl}), so that deficit is real too. That pattern is the point: SFT alone buys strong single-turn instruction following and a viable reasoning initialization, and the deficits it leaves---GPQA and multi-turn following---are exactly what the RL stages target. Across the RL pipeline, GPQA pass@1 rises from 68.18 to 75.6 and IFEval from 88.33 to 95.4 (Table~\ref{tab:sft_results} to Table~\ref{tab:main_comparison}).

\FloatBarrier
\subsection{Reasoning RL}
\label{sec:reasoning-rlvr}

Reasoning RL applies reinforcement learning with verifiable rewards to the SFT-initialized policy across math reasoning, algorithmic puzzle-solving, and scientific QA, and it runs first among the RL stages (\S\ref{sec:pipeline}). The main lesson is that, once the SFT model is strong enough, progress depends more on data curation than on a novel RL objective: the choices that mattered most were keeping the prompt set in the productive learning band, verifying that each prompt admits at least one valid solution path, and not letting rollout budget go to needless verbosity.

\paragraph{Data.}
\label{sec:rlvr-data}
We assemble the reasoning prompt set from three streams: public reasoning problems with reference solutions
(primarily crawled from open repositories); synthesized and augmented data, including synthesized math and logic puzzles generated through
Enigmata~\cite{bytedance2025enigmata} and ReasoningGym~\cite{reasoninggym2025} with task-specific
verification functions; and hard third-party math sets with reference answers. Every problem is
auto-tagged into fine-grained domains, deduplicated, and screened for contamination and policy issues.
The contamination screen is the same one applied to the SFT mix (\S\ref{sec:sft}): the chained n-gram
passes over the reported benchmarks, plus the exhaustive n-gram and dense-retrieval detectors for the
math and knowledge benchmarks that are most prone to leakage.
The assembled prompt set spans three task families of verifiable single-turn prompts, summarized in
Table~\ref{tab:rlvr-data}. The mix is deliberately skewed: \emph{Math} dominates prompt counts, while
\emph{Puzzles} dominate prompt tokens ($\sim$688 tokens/prompt), reflecting long
serialized boards, grids, and rule specifications from 134 task generators in 15 reasoning categories. Two filters then gate entry into RL, and together they are the most important choice in this
stage. A \emph{correctness} filter removes prompts for which a strong teacher (GPT-OSS-120B) obtains no positive
reward. This removes mis-specified or effectively unsolved tasks and provides evidence that each retained prompt admits at least one valid solution path. A \emph{learnability} filter~\cite{foster2025lilo} then keeps prompts in the \emph{productive
learning band} for the current policy: prompts solved at a rate above 0.8 are
dropped as too easy, and prompts with zero observed success are dropped as currently unlearnable. What
remains gives useful gradient and, as the policy improves, forms an automatic curriculum (the online
form is described below). Rewards remain cheap to verify throughout: 83.5\% of
prompts request a \texttt{\textbackslash boxed\{\}} answer matched against a short gold label (median
5 tokens).

\begin{table}[!ht]
\centering
\small
\setlength{\tabcolsep}{6pt}
\caption{\textbf{RLVR prompt set.} Sources, verifiers, and composition per task family.
\emph{Math} leads on prompt count, \emph{Puzzles} on token volume---puzzle statements are roughly
six times longer. Families are assigned by a keyword heuristic because the crawled problems carry no
subject labels. We apply difficulty and correctness filtering (see text) uniformly across families.}
\label{tab:rlvr-data}
\vspace{-0.9em}
\begin{tabular}{@{}lllrrrr@{}}
\toprule
& & & \multicolumn{2}{c}{\textbf{Prompts}} & \multicolumn{2}{c}{\textbf{Prompt tokens}} \\
\cmidrule(l){4-5}\cmidrule(l){6-7}
\textbf{Task} & \textbf{Data source} & \textbf{Verifier} & \textbf{\#} & \textbf{\%} & \textbf{\#} & \textbf{\%} \\
\midrule
Math    & HF crawl; synthesized  & Math-Verify (canonical match) & 57{,}736 & 47.7 &  6.34M & 25.3 \\
Science & HF multi-science crawl & Fuzzy string matching         & 43{,}199 & 35.7 &  4.81M & 19.2 \\
Puzzles & Enigmata; ReasoningGym & Generated Python checkers     & 20{,}226 & 16.7 & 13.91M & 55.5 \\
\midrule
\textbf{Total} & & & \textbf{121{,}161} & \textbf{100} & \textbf{25.07M} & \textbf{100} \\
\bottomrule
\end{tabular}
\end{table}

% ------------------------------------------------------------
\paragraph{Reward design.}
\label{sec:rlvr-algorithm}
Deterministic verifiers deliver the rewards: Math-Verify~\cite{mathverify2024} on canonicalized final answers for math, generated Python checkers executed against the model output for logic puzzles, and fuzzy string matching against reference answers for science QA. In every case the reward is binary, low-noise, and programmatically auditable, which is why this stage runs early in the sequence.

\paragraph{Training setup.}
We use Group Sequence Policy Optimization (GSPO;~\cite{zheng2025gspo}) as the policy-gradient backbone, a practical rather than algorithmic choice: GSPO defines the importance ratio and clipping at the \emph{sequence} level, which is more stable on long MoE rollouts and, together with Rollout Routing Replay, absorbs the mismatch between the FP8 rollout and BF16 training. We refer the reader to the original paper for the objective.
Off-policy correction across the two optimizer steps per rollout uses truncated importance sampling.
The clip range is tight, $\varepsilon = 10^{-3}$ and $\varepsilon_{\mathrm{high}} = 2\times10^{-3}$; KL and entropy coefficients are zero.
The optimizer is AdamW with $\mathrm{lr} = 10^{-6}$ constant, weight decay~$0.1$, $(\beta_1, \beta_2) = (0.9, 0.98)$, and gradient clipping at $1.0$.
Each rollout samples 256 prompts with $n{=}16$ samples per prompt at temperature $1.0$ under a 30{,}000-token response budget, and feeds two optimizer steps (global batch 2{,}048 sequences).
We mask the loss on truncated samples, whose rate grows from $\sim$3\% early in training to 15--20\% after $\sim$40 rollouts as responses lengthen.
Training runs synchronously on 8~nodes ($64\times$H200), with rollouts served by colocated FP8-quantized SGLang engines (8~GPUs per engine) on the same GPUs that run BF16 training, alternating via engine offload.

Two practical controls mattered in this stage. First, we penalize length relative to the shortest correct rollout in each group---a linear penalty that is zero at or below that baseline and grows to a cap at the maximum allowed length, related to but not the same as the group-relative shaping in Kimi~k1.5~\cite{kimi2025k15}. This is partly a behavioral preference (RL-trained reasoning models tend to inflate length without commensurate accuracy gains~\cite{chen2025overthinking}) and partly training economics: under a fixed RL budget, unchecked verbosity cuts rollout throughput and raises truncation. Second, we filter prompts online by group-average reward $\bar r$, following the dynamic-sampling idea in DAPO~\cite{dapo2025}: the selector over-samples $4\times$ (1{,}024 prompts) and keeps only groups with $\bar r \in (0, 0.8]$, dropping all-fail groups (no signal) and near-all-pass groups (negligible gradient). Because the window is fixed but the policy improves, previously dead prompts keep crossing into the productive band, so the difficulty frontier advances on its own without manual staging.

\paragraph{Result anchor.}
Relative to the SFT checkpoint it trains from, the Reasoning RL checkpoint raises GPQA by $+5.3$ points, from 68.2 to 73.5, and leaves both AIME years 2.6--2.8 points lower (Table~\ref{tab:reasoning_rl_results}). We tuned the stage against GPQA, which starts at 68, and treated AIME as a guardrail on which a drop within evaluation noise was acceptable: the SFT base already sits at 90--91 pass@1, and with 30 problems per year a change of under three points is less than one problem's worth (same-condition re-runs of AIME~25 on the SFT checkpoint differ by 4.6 points at $n{=}8$). The takeaway we would carry forward is that, on a strong SFT base, verifier-based Reasoning RL is reliable and largely curation-bound: the gains track the quality of learnability and correctness filtering more than the details of the policy-gradient objective.

\begin{table}[!t]
\centering
\small
\setlength{\tabcolsep}{7pt}
\caption{\textbf{Reasoning RL results}, measured against the SFT checkpoint the stage trains from (SFT step~3799). $\Delta$ is the change contributed by this stage. All values are pass@1 percentages under the same evaluation protocol. GPQA is the stage's primary reference metric; AIME~25 and AIME~26 are guardrails.}
\label{tab:reasoning_rl_results}
\vspace{-0.9em}
% Source (2026-09-15): xth's screenshot of the 3799-lineage eval table, columns
% "SFT 3799 pass@1" and "RL v4-023 pass@1" (AIME25 90.80 / 88.02, AIME26 90.00 / 87.40).
% Only these two benchmarks were in that table. The previous version of this table was
% the 6599 lineage (Reasoning RL Experiments Tracker, quip bAbRAPnDxlak, block "Using the
% step 6599 SFT checkpoint", iter_0000039_hf: GPQA 68.40->75.25, AIME25 86.50->87.50,
% AIME26 85.70->88.23, LCB 61.71->70.71), kept here for reference.
\begin{tabular}{lccc}
\toprule
\textbf{Checkpoint} & \textbf{GPQA} & \textbf{AIME~25} & \textbf{AIME~26} \\
\midrule
Rufus-Air SFT & 68.18 & 90.83 & 90.00 \\
$+$ Reasoning RL & 73.50 & 88.02 & 87.40 \\
\addlinespace[2pt]
\midrule
$\Delta$ & $+5.32$ & $-2.81$ & $-2.60$ \\
\bottomrule
\end{tabular}
% Transposed to checkpoints-as-rows so this table matches tab:coding_rlvr_results,
% tab:if-rl, tab:general-agent-rl and tab:rlhf-results, with the delta as a bottom row.
% LiveCodeBench, Frontier Science, AIME 24, HMMT and Arena-Hard are not shown: no
% values measured on BOTH endpoints of this checkpoint pair (SFT 3799 / RL v4-023) are
% available. Do not fill either endpoint from a different checkpoint to complete a column.
\end{table}

\FloatBarrier
\subsection{Coding RL}
\label{sec:coding_rlvr}

Coding RL builds on the Reasoning RL checkpoint and improves competitive-programming performance with execution-based rewards.

\paragraph{Data composition.}
The Coding RL problem set aggregates four sources (Table~\ref{tab:coding_rlvr_data}) and mixes two
verification modalities: \emph{functional} problems carry \verb|assert| unit tests that exercise a
named function, and \emph{stdin/stdout} problems carry input/output string pairs. A
modality-specific system prompt fixes the answer format: Python inside a fenced
\verb|```python ... ```| block.

The sources differ in kind, not only in size. EvolveCoder~\cite{ruan2026evolvecoder} supplies rewritten problems with adversarially evolved test cases, derived from TACO~\cite{li2023taco}, APPS~\cite{hendrycks2021apps}, Codeforces, and other open collections, and accounts for nearly all functional problems in our set; Nemotron~\cite{ahmad2025opencodereasoning} is real
contest material, Codeforces-dominant with AtCoder, Aizu Online Judge, HackerEarth, CodeChef,
Kattis, and HackerRank in the tail; Dolci~\cite{teamolmo2025olmo3} is the only mixed source; and
ADR~\cite{zheng2026adr} contributes technique-tagged algorithmic problems (greedy, dynamic
programming, sorting). Aggregated across sources, Codeforces is the largest single platform
($\sim$8{,}700 problems, ${\sim}30\%$), so the set stays contest-heavy while spanning both answer
formats.

\begin{table}[!ht]
\centering
\small
\setlength{\tabcolsep}{8pt}
\caption{Coding RL data composition after the drop-all-pass screen. All problems are Python.
\textsuperscript{$\dagger$}Dolci splits into 911 functional and 2{,}198
stdin/stdout problems.}
\label{tab:coding_rlvr_data}
\vspace{-0.9em}
\begin{tabular}{@{}llrr@{}}
\toprule
\textbf{Source} & \textbf{Modality} & \textbf{Problems} & \textbf{\% of set} \\
\midrule
EvolveCoder (synthetic)   & functional              & 14{,}743 & 50.1\% \\
Nemotron (competitive)    & stdin/stdout            &  9{,}393 & 31.9\% \\
Dolci                     & mixed\textsuperscript{$\dagger$} &  3{,}109 & 10.6\% \\
ADR (algorithmic)         & stdin/stdout            &  2{,}160 &  7.3\% \\
\midrule
\textbf{Total}            & \textbf{53.2\% fn / 46.8\% i/o} & \textbf{29{,}405} & \textbf{100\%} \\
\bottomrule
\end{tabular}
\end{table}

\paragraph{Data preprocessing.}
We apply three filters at assembly time and two sampling-based screens before training begins.
\emph{Schema validation} drops records without a non-empty test payload --- an \verb|assert| list for functional problems, \texttt{inputs}/\texttt{outputs} pairs for stdin/stdout ones --- since problems without tests provide no RLVR signal.
\emph{Deduplication} keys on the SHA-256 of the whitespace-normalized, lowercased problem statement.
We check \emph{eval leakage} against LiveCodeBench~v6 (\texttt{release\_v6}, 1,055 problems, 2023-05 to 2025-04) using exact matching of 400-character prefix hashes and character-level 60-gram overlap. Among the 1,055 evaluation problems, \textbf{none has a matching prefix hash or shares at least five character-level 60-grams with any training problem}.
Our provenance checks identify no LeetCode-sourced training problems and place the identified AtCoder problems before the LCB~v6 AtCoder evaluation window. These checks complement the text-overlap screening but do not exclude rewritten or semantically equivalent problems.
A \emph{teacher-based solvability} screen mirrors the reasoning stage (\S\ref{sec:rlvr-data}): we retain only problems for which at least one sampled completion from GPT-OSS-120B passes the verifier, providing an empirical check of problem solvability and test consistency before policy sampling.
The \emph{difficulty} filter is the ``drop-all-pass'' screen that produces the final problem set of Table~\ref{tab:coding_rlvr_data}: before RL begins, we draw four warm-up samples per problem from the base policy and remove every problem solved by all four (pass rate $>0.8$ at $n{=}4$).
This heuristic reduces the prevalence of easy problems, since rollout groups with identical rewards provide no reward-driven policy-gradient signal under the group-mean baseline. Passing all four screening samples, however, does not imply that a problem will remain saturated in subsequent rollouts.
We retain problems with no passing warm-up sample, as they may yield successful completions and informative reward variation in subsequent rollouts.
We do not repeat the difficulty filter during training.

\paragraph{Reward design.}
We parse each rollout response for the last fenced \verb|```python| code block and execute it in a sandboxed Python runtime against the problem's tests, dispatching on the label modality: \verb|assert|-based unit tests for functional problems, stdin/stdout comparison for the rest.

Per-problem tests are sub-sampled to at most 50 with a deterministic, SHA-256-seeded random choice, so the sub-sample is identical across rollouts of the same problem and the reward does not drift with sampler randomness.

The reward is binary: 1 if all selected tests pass for a given completion, 0 otherwise.
We apply no additional format penalty, as fenced-block compliance is already 98--100\% at iteration~0.

\begin{figure}[!t]
    \centering
    % Drawn at its final on-page size (plot_coding_rlvr_curves.py) --- include
    % WITHOUT [width=...] so the in-figure text keeps its true point size.
    \includegraphics[width=0.70\linewidth]{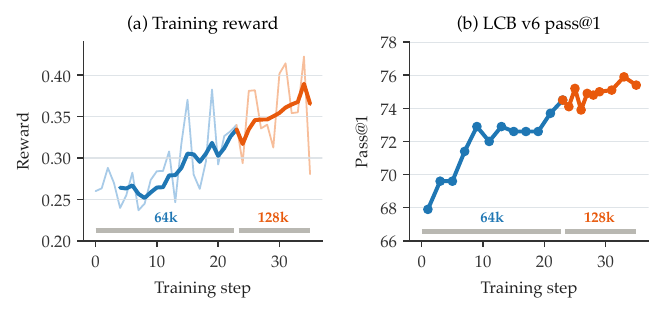}\vspace{-10pt}
    \caption{Coding RL training dynamics. (a)~Raw training reward. (b)~LiveCodeBench~v6 pass@1. The maximum rollout response length is 64K tokens for the first 23 rollout steps and 128K thereafter. Following the extension, the truncation rate falls from 11--28\% to below $0.1\%$. During continued training, reward reaches 0.42 and LiveCodeBench~v6 pass@1 peaks at 75.9 at step~33, the checkpoint we carry forward.}
    \label{fig:coding_rlvr_curves}
    % Back-compat aliases: existing \ref{fig:coding_rlvr_reward} and
    % \ref{fig:coding_rlvr_lcb} in the prose resolve to this figure's number.
    % (Optional polish: switch those refs to \ref{fig:coding_rlvr_curves}(a)/(b)
    % for panel-precise pointers, then delete these two aliases.)
    \label{fig:coding_rlvr_reward}\label{fig:coding_rlvr_lcb}
\end{figure}

\paragraph{Training setup.}
We optimize the policy with GSPO and apply truncated importance sampling to mitigate the mismatch between FP8 rollout inference and BF16 training.
The clip range is tight, $\varepsilon = 2\times10^{-3}$ and $\varepsilon_{\mathrm{high}} = 3\times10^{-3}$, matching the sequence-level GSPO loss.
KL and entropy coefficients are zero for most of training; we enable a nominal KL term ($10^{-6}$) in the 128K continuation as a stability guard.
The optimizer is AdamW with $\mathrm{lr} = 10^{-6}$ constant, weight decay~$0.1$, $(\beta_1, \beta_2) = (0.9, 0.98)$, and gradient clipping at $1.0$.
Each rollout consists of 128 prompts with $n{=}64$ samples per prompt (8{,}192 sequences; 64 prompts after the 128K extension), followed by eight optimizer steps. We mask the loss on responses truncated at the generation limit; these account for less than $0.1\%$ of samples after the extension to 128K tokens.
As shown in Figure~\ref{fig:coding_rlvr_reward}, the raw training reward climbs from 0.26 to 0.34 over the first 23 rollout steps under a 64K-token response budget.
Response lengths grow throughout this phase, and 11--28\% of samples per rollout hit the 64K cap and are loss-masked, so at step 23 we extend the maximum response length to 128K and resume from that checkpoint.
The extension effectively eliminates truncation ($<$0.1\% of samples thereafter) and reward resumes its climb, reaching 0.42 by step 34.
Figure~\ref{fig:coding_rlvr_lcb} shows the corresponding LiveCodeBench~v6 pass@1, which rises from 67.9 at step 1 to 74.5 across the 64K phase and peaks at 75.9 at step 33 under the extended budget, the checkpoint we carry forward.
Training runs on 32 eight-GPU nodes (256 GPUs total) in a colocated configuration: FP8-quantized SGLang engines serve rollouts on the same GPUs that run BF16 training, alternating via engine offload.

% \begin{figure}[t]
%     \centering
%     \begin{subfigure}[t]{0.4\linewidth}
%         \centering
%         \includegraphics[width=\linewidth]{materials/coding-rlvr-reward.pdf}
%         \caption{}
%         \label{fig:coding_rlvr_reward}
%     \end{subfigure}
%     % \hfill
%     \begin{subfigure}[t]{0.4\linewidth}
%         \centering
%         \includegraphics[width=\linewidth]{materials/coding-rlvr-lcb.pdf}
%         \caption{}
%         \label{fig:coding_rlvr_lcb}
%     \end{subfigure}
%     \caption{Coding RL training dynamics. (a)~Raw training reward: light curves show
%     per-step reward and dark curves a 5-step moving average. (b)~LiveCodeBench~v6 pass@1
%     of intermediate checkpoints. The maximum response length, indicated at the bottom of
%     each panel, is 64k tokens for the first 23 rollout steps and 128k thereafter:
%     11--28\% of samples per rollout hit the 64k cap during the first phase, and extending
%     the budget at step~23 removes truncation almost entirely ($<$0.1\%), lifting reward
%     to 0.42 and pass@1 to its peak of 75.9 at step~33, the checkpoint we carry forward.}
%     \label{fig:coding_rlvr_curves}
% \end{figure}

\paragraph{Result anchor.}
Relative to the Reasoning RL checkpoint, Coding RL lifts LiveCodeBench~v6 pass@1 by $+7.3$ points (Table~\ref{tab:coding_rlvr_results}), with the full trajectory in Figure~\ref{fig:coding_rlvr_lcb}: most of the gain accrues in the 64K phase, and pass@1 peaks at step~33 after the response-budget extension.
Pass@8 improves by $2.3$ points, compared with $7.3$ points for pass@1 (Table~\ref{tab:coding_rlvr_results}).
The practical finding from this run is that extending the response budget removed truncation masking, after which both reward and LiveCodeBench kept rising.

\begin{table}[!t]
\centering
\small
\setlength{\tabcolsep}{25pt}
\caption{\textbf{Coding RL results} on LiveCodeBench~v6, measured against the Reasoning RL checkpoint the stage trains from. $\Delta$ is the change contributed by this stage. All values are percentages from one single-stage evaluation at $n{=}8$; this protocol differs from the one behind Table~\ref{tab:main_comparison}, so the LiveCodeBench values here are not directly comparable with the headline table.}
\label{tab:coding_rlvr_results}
\vspace{-0.9em}
\begin{tabular}{lcc}
\toprule
\textbf{Checkpoint} & \textbf{Pass@1} & \textbf{Pass@8} \\
\midrule
Reasoning RL & 68.6 & 85.1 \\
$+$ Coding RL & 75.9 & 87.4 \\
\addlinespace[2pt]
\midrule
$\Delta$ & $+7.3$ & $+2.3$ \\
\bottomrule
\end{tabular}
\end{table}

% [IMPORTANT NOTE FROM ZHIHAN: the following Instruction-Following RL section has been polished and audited by Zhihan himself, so please leave the section as it is. No further modifications or adjustments are needed.]

\FloatBarrier
\subsection{Instruction-Following RL}
\label{sec:if-rl}

Instruction-following (IF) RL targets the capability of following complex user instructions precisely. We decompose ``complex'' along two axes and build training data for each. The first is \emph{multi-constraint} instructions, where a single request carries several constraints that must all be satisfied at once; we concentrate on rule-verifiable hard constraints such as length limits, structural templates, and keyword requirements. The second is \emph{multi-turn} instruction following, which requires retaining layered instructions across conversation turns, adhering to the system prompt for the whole dialogue, and staying self-coherent over extended exchanges. We synthesize one dataset per axis and train them jointly in a single stage.

\paragraph{Verifiable multi-constraint IF data.}
We synthesize 14K single-turn prompts whose constraints can all be checked by Python. We start from a manually written pool of single-constraint instructions, then have a teacher LLM (Qwen3-235B-A22B) automatically augment it into diverse multi-constraint instructions carrying 2--6 atomic constraints each. An example could be ``answer in exactly three paragraphs, keep every paragraph under 40 words, and include the word `therefore' exactly once''. The teacher model also generates a Python verification function per constraint. We combine these instructions with an existing set of user queries to form training prompts, then apply two filters: difficulty filtering removes prompts that are too easy for the policy model ($>$80\% pass rate across 8 sampled responses), and quality filtering removes prompts where the teacher model itself cannot produce a passing response in 4 attempts. Each atomic constraint becomes one rubric, and its Python function decides whether that rubric is satisfied.

\begin{figure}[!t]
\centering
\includegraphics[width=0.98\textwidth]{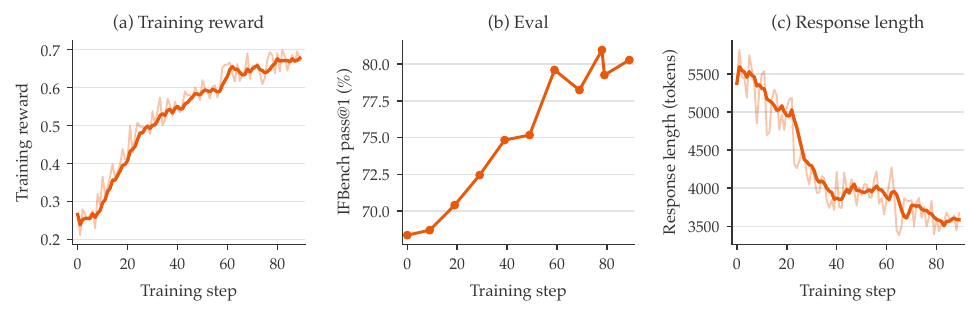}
\vspace{-6pt}
\caption{\textbf{IF RL training dynamics}. In (a) and (c) the light curve is the per-step value and the dark curve a 5-step moving average. Reward rises throughout without flattening, so the run ends on its step budget rather than on convergence. Response length falls from ${\sim}5.4$K tokens to ${\sim}3.6$K and stays there. The IFBench eval in panel~(b) is scored by the in-loop training-time harness, so its values differ slightly from the reported protocol of Table~\ref{tab:if-rl}.}
\label{fig:ifrl_curves}
\end{figure}

\nocite{qubric}

\paragraph{Multi-turn conversational IF.}
We synthesize multi-turn conversations and train the model to produce the response to the final user turn, conditioned on the preceding dialogue. The conversations are generated adversarially: a teacher LLM plays the user, while several open-weight models take the assistant role in rotation across data points, so that the training data are not biased toward a single model's style. The user-side model also plans how the conversation unfolds and actively tries to make the assistant fail, such as layering instructions, revising earlier requirements, and introducing distractors. These yield hard instruction-following instances that current mainstream models do not handle reliably. A portion of the data carries a system prompt and serves as instruction-hierarchy training~\cite{wallace2024hierarchy, IHEval}: the model must keep honoring the system prompt's requirements throughout the dialogue, including when the user turns challenge the system prompt. Each conversation is paired with a list of rubrics that evaluate the response to the final turn. Rubrics that can be verified programmatically are checked by Python functions; the rest are left to the LLM judge (e.g., ``Does the response avoid recommending dishes containing shellfish, given the user's allergy stated in turn 1?'').

We apply three filtering stages: (i) \emph{sanity filtering} removes overlong conversations caused by verbose assistant responses and downsamples cases where the expected behavior is to decline the user's request, to prevent over-refusal; (ii) \emph{quality filtering} uses the teacher model to attempt each conversation---if it cannot pass the rubrics at least once in 4 attempts, we drop the sample as likely unanswerable; (iii) \emph{difficulty filtering} removes cases that are too easy for the policy model ($>$80\% pass rate across 8 sampled responses). The final dataset contains 13K conversations averaging 6.1 turns and 2.7 rubrics each.

\paragraph{Reward design.}
Both datasets share one reward formulation: a \emph{rubric-based binary reward}. For each training instance we express the conditions the response must meet as a list of rubrics, score every rubric independently, and grant positive reward only when \emph{all} of them are satisfied. Rubrics come in two forms. \emph{Code-verifiable} rubrics are checked by a Python function that returns true or false---the natural choice for mechanical conditions such as paragraph counts, casing, or forbidden words. \emph{LLM judge} rubrics are scored by a teacher LLM and cover the conditions no program can check, such as whether the response respects a preference the user stated several turns earlier. Crucially, the rubrics of an instance describe only what the response \textbf{must} satisfy. We deliberately exclude ``optional'' rubrics---desirable but not required properties---because they inject noise into the reward signal and push the policy toward a length bias.

\begin{table}[!t]
\centering
\small
\caption{\textbf{Instruction-following RL results}, measured against the Coding RL checkpoint the stage trains from. $\Delta$ is the change contributed by this stage. IFEval, IFBench and AdvancedIF report prompt-level accuracy.}
\label{tab:if-rl}
\vspace{-0.9em}
\resizebox{\textwidth}{!}{%
\begin{tabular}{lcccccc}
\toprule
\textbf{Checkpoint} & \textbf{IFEval} & \textbf{IFBench} & \textbf{Multi-challenge} & \textbf{AdvancedIF} & \textbf{GPQA} & \textbf{AIME~25} \\
\midrule
Coding RL & 90.5 & 63.8 & 31.1 & 45.5 & 67.9 & 87.6\\
$+$ IF RL & 94.5 & 77.8 & 55.8 & 62.9 & 71.1 & 86.4 \\
\addlinespace[2pt]
\midrule
$\Delta$ & $+4.0$ & $+14.0$ & $+24.7$ & $+17.4$ & $+3.2$ & $-1.2$ \\
\bottomrule
\end{tabular}}
\end{table}

\paragraph{Training setup.}
We mix the two datasets and train them jointly in a single stage on top of the Coding RL checkpoint. We train with Group Relative Policy Optimization (GRPO;~\cite{shao2024deepseekmath}): each rollout consists of 256 prompts with $n{=}16$ samples per prompt, the response budget is 16K tokens, and the learning rate is $1.5\times10^{-6}$. We train for 90 steps and carry the final checkpoint forward.

\paragraph{Result anchor.}
Table~\ref{tab:if-rl} measures the stage against the Coding RL checkpoint it trains from, and the gains land on the capabilities the data targets. On hard formatting constraints, IFEval rises $+4.0$ points to 94.5 and IFBench $+14.0$ points to 77.8. On multi-turn instruction following, Multi-challenge rises $+24.7$ points to 55.8. AdvancedIF~\cite{he2025advancedif}, which covers multi-constraint instructions, multi-turn instructions, and system-prompt following, rises $+17.4$ points to 62.9.

Three observations from this stage are worth recording. (i) \emph{IF RL does not cost reasoning ability.} GPQA ($+3.2$) and AIME~25 ($-1.2$) stay level or better; neither move reads as a regression. (ii) \emph{Reasoning first is a natural curriculum for IF.} Our initial checkpoint has already been through Reasoning RL, and training on it yields larger IF gains than running the same stage directly on the SFT checkpoint, especially on AdvancedIF whose instructions are the most complex in our evaluation set. Satisfying complex instructions requires the model to reason about the constraints before answering, so the reasoning stage supplies a prerequisite. (iii) \emph{Rubrics scoped to necessary conditions shorten the final response.} Under our setup, the model's final response (after \texttt{</think>}) becomes shorter over training (Figure~\ref{fig:ifrl_curves}), a direct consequence of applying rubrics only to what the response must contain: the policy converges on precise instruction following as its objective and learns that padding beyond it sometimes actively breaks a constraint. Breaking that principle reverses the effect. When we attached rubrics to optional content as well---on the order of ten rubrics per instance---the policy acquired a pronounced length bias, emitting long, redundant responses that try to cover as many possible rubrics as it can.

\FloatBarrier
\subsection{General Agent}
\label{sec:general_agent}

General Agent training runs first among the agent stages because it teaches the skill the others assume: \textbf{general tool use}. It builds a broad, transferable habit---sequence a multi-step plan, emit well-formed tool calls, read the results, recover from errors---that the more specialized coding and search stages then sharpen for their own domains.

\paragraph{Data.} AgentWorldModel (AWM)~\cite{wang2026agent} provides 10K single-server Model Context Protocol (MCP)~\cite{anthropic2024mcp}
tasks distributed across 1K synthetic environments. We apply two filters before
training. 
(i) \emph{correctness}: we discard tasks on which neither a strong
teacher policy (Qwen3-235B-A22B) nor the initial policy ever passes the binary rule-based reward across sampled rollouts. This provides evidence that
every retained task admits a solution path.
(ii) \emph{difficulty}: tasks that the current policy already
solves reliably or never solves (no gradient signal) are
filtered out for training efficiency.  The two
filters together leave $\sim$2K tasks for RL training.

\paragraph{Environment and tools.} We adopt AWM as our foundation and extend it with
in-house modifications to the task set, server backends, and verifier
robustness needed for stable large-scale RL training. Each task is bound to one environment: a scenario-specific database whose schema is derived from that environment's tasks, and an MCP server whose tools read and write that database, about 35 per environment in AWM. A tool call therefore changes the environment state, and the task's verifier checks that state once the episode ends. In one AWM environment about music streaming, for example, a task asks the agent to create a playlist of an artist's most popular tracks, and the environment exposes a synthesized MCP tool \texttt{create\_playlist} that creates a playlist owned by the current user.

\paragraph{Reward design.} We use AWM's code-based
verification functions as the reward signal, reduced to a binary outcome at termination: $1.0$ if the post-rollout server state satisfies the task's success criteria and $0.0$ otherwise. We use no step-level or LLM judge reward; the per-task verification function is the only
signal the policy is trained against. This drops two of AWM's rewards. AWM uses an LLM judge to absorb environment imperfections that can confound a pure rule-based check; our correctness filter handles this instead. Its format reward targets the protocol through which AWM exposes tools, two meta-tools \texttt{list\_tools} and \texttt{call\_tool} that the agent uses to discover a tool and then invoke it; we instead register every tool schema directly in the system prompt, so the agent calls tools by name, with no discovery step to enforce.

\begin{figure}[!t]
\centering
\includegraphics[width=0.98\textwidth]{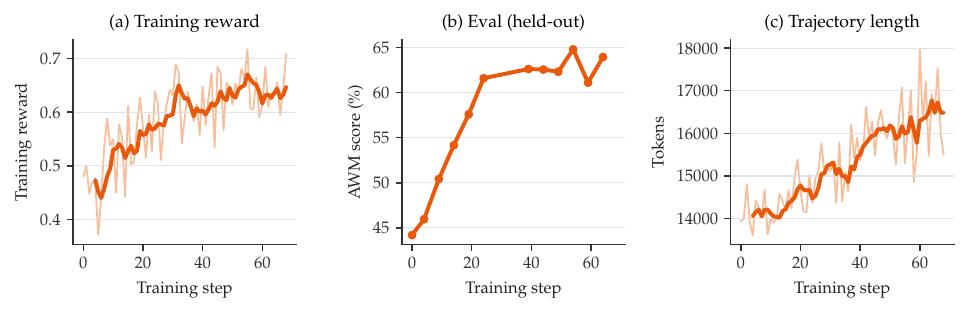}
\vspace{-6pt}
\caption{\textbf{General Agent training dynamics}, shown up to step~68, the checkpoint the pipeline carries forward. (a)~Training reward; (b)~held-out AWM score, scored by the in-loop training-time harness; (c)~trajectory length in tokens. Light curves are per-step values and dark curves a 5-step moving average.}
\label{fig:general_agent_curves}
\end{figure}

\begin{table}[!t]
\centering
\small
\caption{\textbf{General Agent results}, measured against the IF RL checkpoint the stage trains from. $\Delta$ is the change contributed by this stage. Tau2-Retail here is pass@1 under the Claude Sonnet~4.5~\cite{anthropic2025sonnet45} user simulator ($n{=}4$), so it is comparable within this table but not to the Sonnet~5~\cite{anthropic2026sonnet5} row of Table~\ref{tab:main_comparison}.}
\label{tab:general-agent-rl}
\vspace{-0.9em}
% Source: GLM-Air Agentic Training Round 2 Tracker (chorus 5VPK14aPT6XP), rows
% "initial policy (IF, iter_89)" and "AWM_bs48*64_iter68"; updated 2026-09-15.
\begin{tabular}{lcccccc}
\toprule
\textbf{Checkpoint} & \textbf{MCP-Atlas} & \textbf{Tau2-Retail} & \textbf{AIME~26} & \textbf{IFEval} & \textbf{GPQA} & \textbf{LiveCodeBench} \\
\midrule
IF RL & 34.95 & 74.00 & 87.08 & 94.73 & 73.36 & 71.21 \\
$+$ General Agent & 42.75 & 83.80 & 87.60 & 96.16 & 76.39 & 73.36 \\
\addlinespace[2pt]
\midrule
$\Delta$ & $+7.80$ & $+9.80$ & $+0.52$ & $+1.43$ & $+3.03$ & $+2.15$ \\
\bottomrule
\end{tabular}
\end{table}

\paragraph{Training setup.} This stage trains with GRPO and DAPO-style dynamic sampling, which drops prompt groups where all rollouts pass or all fail, and uses Rollout Routing Replay to stabilize MoE training. Rollouts against stateful MCP servers are CPU-heavy (server I/O, database mutation, MCP transport) and run 3{,}072 per training step, so this stage relies on distributed rollout (\S\ref{sec:agent_env}).

\paragraph{Result anchor.} The claim we emphasize is agentic transfer: relative to the checkpoint after IF RL (\emph{Rufus-Air RL}), General Agent training raises MCP-Atlas pass@1 by $+7.80$ points and Tau2-Retail by $+9.80$ points (Table~\ref{tab:general-agent-rl}). We read these gains as evidence that synthetic MCP environments provide a useful general prior for tool orchestration. The non-agentic benchmarks we monitor for regression (AIME~26, IFEval, GPQA, LiveCodeBench) hold or rise modestly. Figure~\ref{fig:general_agent_curves} shows the run behind these numbers: training reward and the held-out AWM score rise together, and the policy reaches them with longer trajectories---from about 14K to about 16.5K tokens over the run.

\FloatBarrier
\subsection{Coding Agent}
\label{sec:coding_agent}

Coding Agent training targets terminal and software engineering tasks. The agent works in a sandbox with a single action, executing a shell command, on a task specified in natural language. Solving a task requires exploring the filesystem, composing shell commands into a multi-step plan, interpreting program and test output, and correcting errors, over episodes of tens to hundreds of tool calls.

\paragraph{Data.} Training tasks are drawn from three public sources spanning terminal and software-engineering work.
\textbf{Endless-Terminal}~\cite{gandhi2026endlessterminals} procedurally generates containerized terminal tasks (file operations, log management, data processing, scripting) with completion tests and no human annotation (MIT).
\textbf{SETA-Env}~\cite{shen2026seta} provides 4{,}500+ verified Harbor-format~\cite{harbor2026} terminal environments, synthesized from web sources such as Ask~Ubuntu and Stack~Overflow and expanded by difficulty-controlled evolution (CC~BY~4.0).
\textbf{Scale-SWE}~\cite{zhao2026scaleswe} supplies executable software-engineering tasks---pre-built Docker images with fail-to-pass unit tests---built from pull requests to public GitHub repositories. This is its executable-task release (CC~BY~4.0); the Scale-SWE-Distilled trajectory corpus of Table~\ref{tab:public_sft_sources} comes from the same collection and shares its GitHub-PR provenance but is a different artifact, distilled SFT trajectories rather than executable tasks.
Preprocessing applies structural filters that remove tasks unsuitable for scaled execution, such as missing required files and multi-container compositions. A learnability filter then removes tasks the current policy solves reliably or never solves, leaving $\sim$4K training tasks; a further $\sim$10K from Scale-SWE remain available but are not used in this run.

\paragraph{Environment and tools.} We adopt the Harbor task format as a unified interface for terminal tasks. Each task is a directory containing an \texttt{instruction.md}, an environment \texttt{Dockerfile}, and a \texttt{tests/test.sh} verifier, so heterogeneous tasks from different sources share one interface. The tool surface is a single \texttt{execute\_command} tool that runs
an arbitrary shell command in the sandbox and returns \texttt{stdout},
\texttt{stderr}, and the exit code. File reads, edits, builds, and test runs are
all expressed as shell commands. We bake each task's container image ahead of time into a dedicated sandbox template with fixed CPU, memory, and storage, so at rollout time the sandbox backend of \S\ref{sec:agent_env} boots the pinned template directly without an online image build, which keeps provisioning fast enough for RL-scale concurrency.

\paragraph{Reward design.} The reward is the task's own verifier, reduced to a binary outcome at termination: the environment uploads \texttt{tests/} into the
sandbox, runs tests against the final state, and returns $1.0$ if the tests all pass and $0.0$ otherwise. No step-level shaping or LLM judge reward is used.

\begin{table}[!t]
\centering
\small
\setlength{\tabcolsep}{4pt}
\caption{\textbf{Coding Agent results}, measured against the General Agent checkpoint the stage trains from. $\Delta$ is the change contributed by this stage. The stage was trained on the compute available rather than to convergence, so these are the values a short run produced.}
\label{tab:coding-agent-rl}
\vspace{-0.9em}
% \begin{tabular}{@{}lccccc@{}}
\begin{tabular}{lccccc}
\toprule
\textbf{Checkpoint} & \textbf{Terminal-Bench 2.1} & \textbf{SWE-bench Verified} & \textbf{AIME~26} & \textbf{IFEval} & \textbf{LiveCodeBench} \\
\midrule
General Agent & 38.76 & 65.60 & 87.60 & 96.16 & 73.36 \\
$+$ Coding Agent & 40.17 & 67.80 & 87.40 & 95.68 & 75.64 \\
\addlinespace[2pt]
\midrule
$\Delta$ & $+1.41$ & $+2.20$ & $-0.20$ & $-0.48$ & $+2.28$ \\
\bottomrule
\end{tabular}
\end{table}

\paragraph{Training setup.}
As in the General Agent stage, we train with GRPO and DAPO-style dynamic sampling, over-sampling to refill the batch after dropping all-pass or all-fail prompt groups, and use Rollout Routing Replay. At this stage's concurrency, some rollouts fail for reasons unrelated to the policy (sandbox boot failure, verifier timeout, dropped connection). We isolate such infrastructure noise: we tag the samples as \textsc{aborted}, distinct from reward-$0$ test failure, and remove them from their sampling group before computing advantages, so they do not bias the group baseline.

\paragraph{Result anchor.} Relative to the General Agent checkpoint, the stage moves SWE-bench Verified pass@1 from $65.60$ to $67.80$ and Terminal-Bench~2.1 from $38.76$ to $40.17$ (Table~\ref{tab:coding-agent-rl}). The non-agentic guardrail benchmarks move by less than half a point (IFEval $-0.48$,
AIME~26 $-0.20$) or rise (LiveCodeBench $+2.28$), so the gain does not
come at their expense.

\emph{Caveat.} This stage was trained on the compute available rather than to convergence. Each reward requires booting a sandbox and running a test suite after an episode of tens to hundreds of tool calls, so a gradient signal costs far more than in Reasoning RL and Coding RL, and the run used only the $\sim$4K tasks described above. The numbers show that the environment, reward, and rollout path work end to end and move the targeted benchmarks, not what the stage converges to. The SWE-bench gain does not survive to the shipped checkpoint, which reads 65.6, this stage's starting value, while Terminal-Bench~2.1 rises further to 42.7 (Table~\ref{tab:main_comparison}). Neither benchmark was measured after Search Agent, so we cannot say which of the two later stages moved them.

\FloatBarrier
\subsection{Search Agent}
\label{sec:deep_research}

% \noindent\emph{Status: before/after reported on three benchmarks (Table~\ref{tab:search-agent-rl}) plus one transferable ablation (context management); the head-to-head comparison against external models on these three benchmarks is in Table~\ref{tab:main_comparison}, measured under this stage's harness.}

The \textbf{Search Agent} works against the open web rather than a bounded, structured tool set: it answers multi-hop factual questions by decomposing them into sub-queries, iterating web search and page extraction, and synthesizing the evidence into a final answer, deciding for itself when it has gathered enough. Unlike the other agent stages, its reward is an LLM judge and therefore noisy; and, as in Coding Agent, a single rollout can run to 100 tool calls, so each unit of gradient signal is expensive. The three choices that mattered most in our runs follow from that: data filtering, reward design, and the optimization algorithm.

\paragraph{Data.}
We extract verifiable question--answer pairs from MiroVerse~\cite{miromind2025mirothinker}, dropping questions whose reference solutions require vision and excluding source subsets that a proxy policy almost never solves, which yields 36{,}614 candidates with 4{,}069 more held out for validation. Because rollouts are expensive, we narrow the candidate set aggressively in two passes. The first pass removes questions the policy answers correctly in any of 8 rollouts \emph{with tools disabled}: such questions are answerable from parametric knowledge alone, and training on them rewards tool-free shortcuts rather than research behavior; this pass removes roughly a quarter of the set. The second pass rolls out the initial policy 8 times per question in the full tool environment, grades each rollout with the same Qwen3-32B judge used as the RL reward (see reward design below), and keeps questions by the raw correct count $c$ out of 8. The two extremes dominate: 15.5\% of questions are never solved and 31.4\% are always solved, and at either extreme all 8 rollouts receive the same reward, so the group-relative advantage is zero and the question contributes no gradient. Keeping $0 < c < 4$ retains questions that are hard but demonstrably solvable, and yields a final RL training set of ${\sim}2.2$k questions.

\paragraph{Environment and tools.}
The agent operates in a three-tool environment:
\begin{itemize}[nosep,leftmargin=*]
    \item \textbf{Web Search}: queries a search backend and returns ranked URLs with snippet text. During RL training we use a lower-cost internal search backend to keep rollout throughput manageable; evaluation uses a separate, standardized tool setup.
    \item \textbf{Web Scrape}: fetches a URL and extracts the main content from the raw HTML. For pages whose extracted content exceeds the model's effective context budget, a summarizer condenses the material into structured evidence and a summary, preserving critical facts (names, dates, numbers) verbatim.
    \item \textbf{Python}: a sandboxed interpreter (no network or filesystem access) for arithmetic, date, and string manipulation.
\end{itemize}
The agent iterates up to 100 tool calls during training and 200 during evaluation, within a 128K-token budget for the whole trajectory (Table~\ref{tab:training_config}). The final answer must be enclosed in \texttt{\textbackslash boxed\{\}}.

\paragraph{Reward design and prompts.}
The reward is outcome-based: an LLM judge scores the final answer against the ground truth on a continuous $[0,1]$ scale, with partial credit for partially correct list- and description-type answers. To reward research behavior rather than parametric recall, we guard the reward against tool-free shortcuts: the judge sees a trajectory only when it commits to an explicit final answer, and positive reward requires at least 2 successful tool calls. The reward is further scaled down in proportion to the fraction of malformed tool calls in the trajectory, down to a factor of $0.5$ in the worst case, and a trajectory that exhausts its budget without producing a final answer receives a penalty of $-0.1$. The reward has no unconditional positive term: an earlier format bonus for producing a boxed answer was removed after the policy learned to collect it with tool-free guesses. The tool call gate verifies that calls executed, not that the answer depends on what they returned, so a memorized answer with incidental tool use remains a residual path that we audit in trajectories rather than close in the reward.
Unlike the deterministic verifiers of Reasoning RL and Coding RL, this channel is noisy: the judge awards partial credit, and repeated grading of the same answer does not always agree; this motivates the advantage estimator below. We use Qwen3-32B as the training judge because it must score 512 trajectories per rollout step; evaluation uses a stronger, independent judge (Appendix~\ref{sec:search_eval_appendix}). The system prompt encodes multi-hop reasoning strategies: question decomposition into independent sub-problems, search reformulation on failure, liberal use of scraping (since snippets alone are insufficient for precision fact-finding), and explicit progress tracking after each research round.

\paragraph{Training setup.}
Like the other agent stages, this stage uses GRPO, here with \emph{mean-only} group advantages, $\widehat{A}_i = r_i - \mathrm{mean}(\{r_j\}_{j=1}^{G})$ with $G=16$, as in Dr.~GRPO~\cite{liu2025drgrpo}. We keep GRPO's per-sequence length normalization, which Dr.~GRPO also removes: on 30--100K-token trajectories a constant normalizer would let the longest trajectories dominate the batch gradient. The reason for dropping the standard deviation normalization is the noisy reward channel. Dividing by the group standard deviation erases the size of the reward gap: for $k$ successes among $G$ rollouts separated from the failures by a gap $\Delta$, the normalized advantage of a success is $\sqrt{(G-k)(G-1)/(kG)}$, independent of $\Delta$, so at $G=16$ a lone success at gap $1.0$ and a lone success at a $0.1$ partial-credit gap both receive $+3.75$, about $3.9\times$ the advantage of a success in a balanced group. The learnability band keeps prompts solved as rarely as one time in eight, so such groups are common, and normalization amplifies grading noise to the scale of a genuine solve. In a paired ablation that differed only in this flag, the mean-only run led on the held-out set at every matched evaluation (55.8 to 64.8 against 49.7 to 58.7), while the normalized run's policy entropy rose faster with no corresponding validation gain. Each rollout of 32 prompts feeds two optimizer steps (global batch 256 sequences), keeping the update nearly on-policy, and Rollout Routing Replay aligns the trainer's expert routing with the rollout engine's. We retain a small KL anchor to the stage's initialization (coefficient $10^{-3}$): without it policy entropy drifted from 0.72 to 1.12, and with it entropy rose gradually to 0.94, alongside validation. We use no entropy bonus; the optimizer is AdamW with $\mathrm{lr} = 3\times10^{-6}$ constant.

\begin{figure}[!t]
\centering
% Rendered by tmp/render_search_agent_curves.py from tmp/mssta3cg_stitched_by_rollout_step (1).csv
% (W&B run mssta3cg stitched by rollout step; columns rollout/raw_reward, eval/search, rollout/tool_iters).
% The step-4 eval point (0.54) is patched in from a reference run: the logged run crashed before its
% first eval finished, so the stitched CSV has no step-4 value (xth, 2026-09-19).
\includegraphics[width=0.98\textwidth]{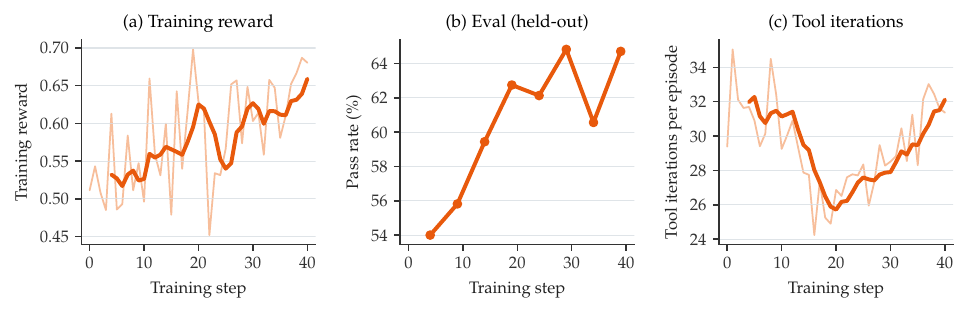}
\vspace{-6pt}
\caption{\textbf{Search Agent training dynamics}. (a)~Training reward, the judge-scored mean over each rollout batch; (b)~pass rate on the held-out 200-prompt validation set, evaluated every five steps (the step-4 point is taken from a reference run, because the logged run crashed before its first evaluation finished); (c)~tool iterations per episode. Light curves are per-step values and dark curves a 5-step moving average.}
\label{fig:search_agent_curves}

\vspace{1.2em}
% Results table kept in the same float as Figure~\ref{fig:search_agent_curves} so the two stay on one page.
{\small
\setlength{\tabcolsep}{7pt}
\captionof{table}{\textbf{Search Agent results} on the three search benchmarks (pass@1 under the standardized evaluation harness: Serper + Jina, scratchpad context management on BrowseComp), measured against the checkpoint the stage trains from. $\Delta$ is the change contributed by this stage. Per-trajectory tool statistics for the same runs are in Table~\ref{tab:tool_patterns}.}
\label{tab:search-agent-rl}
\begin{tabular}{lccc}
\toprule
\textbf{Checkpoint} & \textbf{BrowseComp} & \textbf{Seal-0} & \textbf{HLE-Verified} \\
\midrule
Coding Agent & 34.2 & 48.6 & 47.7 \\
$+$ Search Agent & 37.2 & 54.0 & 51.1 \\
\addlinespace[2pt]
\midrule
$\Delta$ & $+3.0$ & $+5.4$ & $+3.4$ \\
\bottomrule
\end{tabular}
\par}
\end{figure}

\paragraph{Training dynamics.}
Figure~\ref{fig:search_agent_curves} shows the run: training reward and the held-out pass rate rise together over the 40 steps, while tool iterations per episode first fall and then climb back. Per-sample dumps from the run explain the two phases. The fall is compression across the board rather than tool avoidance: winning trajectories (reward $\geq 0.9$) shorten by about 4 calls and 7K tokens at the median and losing ones (reward $\leq 0.05$) by about 7 calls and 16K tokens, and the held-out pass rate keeps rising. The climb is a new tail of productive long searches rather than a return to the earlier behavior: the median winner still makes about 25 calls, but the winners' 90th percentile rises from 40 to 54.5 calls, the number of trajectories with more than 45 calls nearly triples (49 to 133) and their winner share rises from 24\% to 43\%, while winners' text keeps shrinking (42K to 33K tokens at the median). Long searches change from a failure signature into a viable strategy, and the added depth late in the run is spent on searches that succeed rather than on longer trajectories for their own sake.

\paragraph{Result anchor.}
Table~\ref{tab:search-agent-rl} reports the stage's before/after on the three search benchmarks: Search Agent adds $+3.0$ on BrowseComp, $+5.4$ on Seal-0, and $+3.4$ on HLE-Verified over the Coding Agent checkpoint it starts from, with the comparison against other models in Table~\ref{tab:main_comparison}. The search benchmarks are evaluated under a standardized harness whose tools, judge, and scratchpad context management differ from the training setup (Appendix~\ref{sec:search_eval_appendix}); because its search backend is one the policy never used in training, the gains suggest that RL improved general search behavior rather than exploiting the training backend.

\FloatBarrier
\subsection{RLHF}
\label{sec:rm-rl}

Here \emph{RLHF} means RL against a learned reward model, rather than RL directly from human labels---consistent with the no-new-annotation stance of the rest of the recipe. It serves open-ended quality, where no deterministic verifier exists. Because a learned reward is especially vulnerable to reward hacking and drift~\cite{gao2023scaling}, this stage runs last, which shortens the time a gameable reward is under optimization pressure (\S\ref{sec:pipeline}). Two observations motivate this stage. First, despite the shortening during IF RL (\S\ref{sec:if-rl}), responses, excluding the thinking portion, grow longer over SFT and outcome-reward RL without a commensurate increase in usefulness. Second, none of the earlier stages explicitly evaluates the final response for harmful or incoherent content. RLHF therefore targets concise, helpful, and harmless final responses. We use on-policy RL against the reward model rather than an offline preference method such as DPO~\cite{rafailov2023direct}, so the reward signal acts on the policy's own rollouts rather than on a fixed preference dataset.

\paragraph{Reward model and data.}
Skywork-Reward-V2-Qwen3-8B provides the reward~\cite{skywork2025reward}, an open reward model, rather than an in-house or newly human-labeled model. This keeps the stage reproducible and consistent with the recipe's open-data stance. We considered three public preference datasets as candidate prompt sources: Arena Human Preference~\cite{chiang2024chatbotarena}, HelpSteer3~\cite{wang2025helpsteer3preference}, and HH-RLHF~\cite{bai2022training}. Before running the training ablations, we filtered out examples with missing, empty, or otherwise degenerate responses and scored the remaining chosen and rejected responses with the Skywork reward model. Table~\ref{tab:rlhf-data-filter} reports each dataset's original size, the number of valid examples remaining after filtering, and the average reward-model scores and response lengths of its chosen and rejected responses.

\begin{table}[!ht]
\centering
\small
% Restoring the Arena Human Preference row pushed this table 21pt past \textwidth
% (its 1055.1 / 623.7 cells are the widest in the block). Column padding is the only
% thing tightened; no value is changed and the body stays at \small.
\setlength{\tabcolsep}{3pt}
\caption{\textbf{RLHF prompt datasets.} We score each dataset's chosen and rejected responses with the Skywork reward model; \emph{Valid} counts the pairs left after we drop missing, empty, and otherwise degenerate responses. Bradley--Terry~\cite{bradley1952rank} scores carry no shared zero across datasets, so read the score columns within a row rather than as grounds for ranking the three.}
\label{tab:rlhf-data-filter}
\vspace{-0.9em}
\begin{tabular}{@{}lrrcccc@{}}
\toprule
\textbf{Dataset}
& \textbf{Original \#}
& \textbf{Valid \#}
& \textbf{\shortstack{Avg.\ Chosen\\Score}}
& \textbf{\shortstack{Avg.\ Rejected\\Score}}
& \textbf{\shortstack{Avg.\ Chosen\\Tokens}}
& \textbf{\shortstack{Avg.\ Rejected\\Tokens}} \\
\midrule
Arena Human Preference & 84,402 & 53,502 & 17.06 & 12.08 & 1055.1 & 623.7 \\
HelpSteer3 & 38,459 & 29,506 & 9.60 & 2.95 & 439.1 & 373.2 \\
HH-RLHF    & 112,052 & 75,815 & -4.23 & -8.41 & 78.8 & 72.9 \\
\bottomrule
\end{tabular}
\end{table}

The table shows substantial differences among the candidate datasets. Arena Human Preference contains by far the longest responses and receives high positive scores from the reward model for both sides of each pair. HelpSteer3 contains shorter responses and exhibits the largest average score separation between chosen and rejected responses, but its broader and more complex instructions proved difficult for the reward model to evaluate consistently during on-policy training. HH-RLHF contains much shorter responses, and both sides receive negative absolute scores, although the chosen responses still score higher on average than the rejected responses.

The corresponding training ablations produced three negative findings. First, training on the unfiltered HH-RLHF prompt set noticeably degraded other capabilities, most visibly coding and instruction following, motivating the validity filter described above. Second, although HelpSteer3 provides more diverse tasks, its prompts were difficult for the reward model to score reliably: reward increased slowly while scientific reasoning and instruction following regressed. Third, instruction-following and agentic prompts were unsuitable for this stage because the reward model could not reliably assess complex constraint satisfaction. We therefore train the final stage on the \textbf{75{,}815 HH-RLHF prompts} that remain after the validity filter of Table~\ref{tab:rlhf-data-filter}. HH-RLHF's chosen and rejected responses and preference labels serve only the offline dataset analysis summarized in Table~\ref{tab:rlhf-data-filter}; RLHF training uses only the prompts, and the open reward model generates all rewards for the policy's on-policy rollouts. During training, we additionally monitor reward on Arena-Hard~v2~\cite{li2024crowdsourced} as an external diagnostic.

\begin{figure}[!t]
\centering
\includegraphics[width=0.98\textwidth]{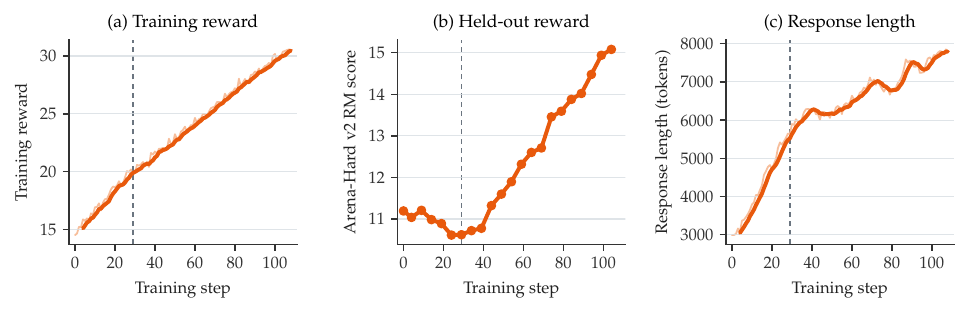}
\vspace{-10pt}
\caption{\textbf{RLHF training dynamics} of the shipped run, trained on HH-RLHF prompts against the Skywork-Reward-V2-Qwen3-8B reward model with a linear length penalty. (a)~Training reward; (b)~held-out reward-model score on Arena-Hard~v2 prompts, evaluated every five steps; (c)~mean response length. Light curves are per-step values and dark curves a 5-step moving average; the dashed line marks iteration~29, the checkpoint shipped as \emph{Rufus-Air}.}
\label{fig:rlhf_curves}
\end{figure}

\paragraph{Reward design.}
This stage differs structurally from Reasoning RL and Coding RL because Bradley--Terry reward scores are unbounded and noisy: a higher pointwise score does not always correspond to a better response. The shipped run (Figure~\ref{fig:rlhf_curves}) optimizes the raw reward-model score under a linear length penalty, which reduces the score of over-length responses when their raw reward is positive while leaving negative rewards unchanged. Optimization uses GRPO.

\begin{table}[!t]
\centering
\small
\setlength{\tabcolsep}{4.8pt}
\caption{\textbf{RLHF results.} Vanilla RLHF optimizes the raw reward-model score under a linear length penalty (\S\ref{sec:rm-rl}); the resulting checkpoint (iteration~29 of the run in Figure~\ref{fig:rlhf_curves}) is the shipped model, \emph{Rufus-Air}. The reference row is the Search Agent checkpoint the stage starts from, so $\Delta$ is the change over the RLHF stage. Both columns are Arena-Hard~v2 win rates.}
\label{tab:rlhf-results}
\vspace{-0.9em}
% Source (2026-09-15): xth's two screenshots of the v6 RLHF run evals
% (GLM-4.5-Air-rmrl-bsz512-lp10k-apl20-iter29-v6, W&B 2w3un5kt).
%   rung r4 : arena 0.6892  a-hard 0.8306  a-creat 0.3856 | ifbench 0.8000 aime24 0.8500 aime25 0.9000
%   rung r29: arena 0.7815  a-hard 0.8905  a-creat 0.5297 | ifbench 0.8067 aime24 0.8375 aime25 0.9000
% "arena" (overall) is not shown because the paper reports only the two splits.
% r29 a-hard/a-creat match the Rufus-Air Arena-Hard cells of tab:main_comparison (89.1/53.0);
% r29 AIME25 (90.00) differs from the main table's 88.3 (different eval run / n).
% Previous version of this table (8-1/8-2 vanilla vs rank on the old lineage) is in git.
\begin{tabular}{lcc}
\toprule

& \textbf{\shortstack{Arena-Hard v2 (HP)}}
& \textbf{\shortstack{Arena-Hard v2 (CW)}} \\
\midrule
Search Agent & 83.06 & 38.56 \\
+ RLHF & 89.05 & 52.97 \\
\addlinespace[2pt]
\midrule
$\Delta$ & $+5.99$ & $+14.41$ \\
\bottomrule
\end{tabular}
\end{table}

\paragraph{Training dynamics.}
Figure~\ref{fig:rlhf_curves} shows the shipped run: training reward (a), the held-out reward-model score on Arena-Hard~v2 prompts (b), and mean response length (c), with iteration~29 marked. Both reward panels report the Skywork reward optimized during training rather than the downstream evaluation scores in Table~\ref{tab:rlhf-results}.

\paragraph{Result anchor.}
Table~\ref{tab:rlhf-results} reports the shipped run against the Search Agent checkpoint it starts from: Arena-Hard~v2 Hard Prompt rises from 83.06 to 89.05 ($+5.99$) and Creative Writing from 38.56 to 52.97 ($+14.41$), and the two Rufus-Air cells are the Arena-Hard values of Table~\ref{tab:main_comparison}. On the monitored benchmarks the run's own evaluations move by at most 1.3 points over the stage: IFBench 80.0 to 80.7, AIME~24 85.0 to 83.75, AIME~25 unchanged at 90.0 (these are the run's evaluations at their own sample count; the AIME~25 cell of Table~\ref{tab:main_comparison} for the same checkpoint, 88.3, comes from a different evaluation run).

\FloatBarrier
\section{Training Infrastructure}\label{sec:train_infra}

Our training stack has two in-house components built on open-source software. \textbf{Rufus-Slime}, our adaptation of Slime~\cite{slime2025}, runs SFT and every RL stage, with Megatron-LM~\cite{shoeybi2019megatron} as the training backend and SGLang~\cite{zheng2024sglang} as the rollout engine (\S\ref{sec:sft_rl}). \textbf{Rufus-Gym}, our agent environment and orchestration layer, extends the stack with multi-turn rollout for the agent stages (\S\ref{sec:agent_env}).

\subsection{SFT and General RL}\label{sec:sft_rl}

\paragraph{SFT.}
SFT uses Slime’s SFT path with packed-sequence attention for sequences up to 128K tokens, accommodating approximately 98\% of the original SFT sample mix without modification. The 1.8\% of samples in the 32K--120K range are bucket-sampled with fewer other samples per batch to bound per-step compute. The production run uses an effective batch size of 4096 and takes approximately 2380 steps and 105 wall-clock hours per epoch. The three-epoch schedule runs for about 13 days on 64 nodes, or roughly 832 node-days, making SFT the largest single expense among the stages we can account for. Training uses BF16 throughout, AdamW~\cite{loshchilov2019adamw}, and standard gradient clipping. Loss is computed only on assistant turns; system prompts, user turns, and tool observations remain in context but are masked from the loss. Per-dataset chat templates allow heterogeneous sources, including thinking SFT data and agentic trajectories, to be mixed within the same batch.

\paragraph{General RL.}
General RL runs on the same stack, adding a rollout loop, a reward path, and a set of stability-critical configuration choices. The choices that mattered:

\begin{itemize}
  % \item \textbf{Training--rollout disaggregation (later stages).} For the RL stages after Reasoning RL, the largest observed reliability improvement came from separating rollout serving from policy training across node groups. Policy training runs on a Megatron-LM node group, SGLang rollout workers run on a separate node group, and the two groups communicate through a parameter broadcast and a rollout queue; rollout:training node ratios are tuned per stage so that neither side idles. Before disaggregation, RL runs were routinely bottlenecked by rollout queue saturation and degraded into unstable loss trajectories; after disaggregation, runs of 30+ rollout steps $\times$ 240 training steps complete without manual intervention. (Reasoning RL instead uses synchronous colocated serving, as described in \S\ref{sec:reasoning-rlvr}.)

  \item \textbf{Large rollout batches.} Group-relative methods such as GRPO and GSPO estimate each response's advantage against the other responses sampled for the same prompt, so the estimate is less noisy with more samples per prompt and more prompts per step. Each rollout step therefore samples 4{,}096--8{,}192 responses: 256 prompts $\times$ 16 samples in Reasoning RL and IF RL, and 128 prompts $\times$ 64 samples in Coding RL before its budget extension (Table~\ref{tab:training_config}).

  \item \textbf{Rollout Routing Replay (R3)} On MoE policies, rollout and training can route the same token to different experts even at identical precision, adding a rollout--training log-probability gap on top of any precision mismatch. R3, available in upstream Slime, caches the routing decisions made during rollout and replays them in the training forward pass, removing this component of the gap.
  
  \item \textbf{Mixed-precision (FP8) rollout with BF16 training.} Reasoning RL and Coding RL run rollouts in mixed precision---the first three transformer layers in BF16 and all remaining layers in FP8---while keeping the parameter update fully in BF16.

  \item \textbf{Truncation-ratio monitoring.} Across all RL runs we monitor the fraction of rollouts truncated at the stage's response budget. In Reasoning RL and Coding RL, truncated samples are masked from the loss, so a rising ratio shrinks the share of each batch that contributes gradient; Coding RL responded by extending its response budget from 64K to 128K mid-run (\S\ref{sec:coding_rlvr}).

\end{itemize}

\subsection{Agentic RL and Its Environment}\label{sec:agent_env}

\begin{figure}[!ht]
\centering
\includegraphics[width=0.8\linewidth]{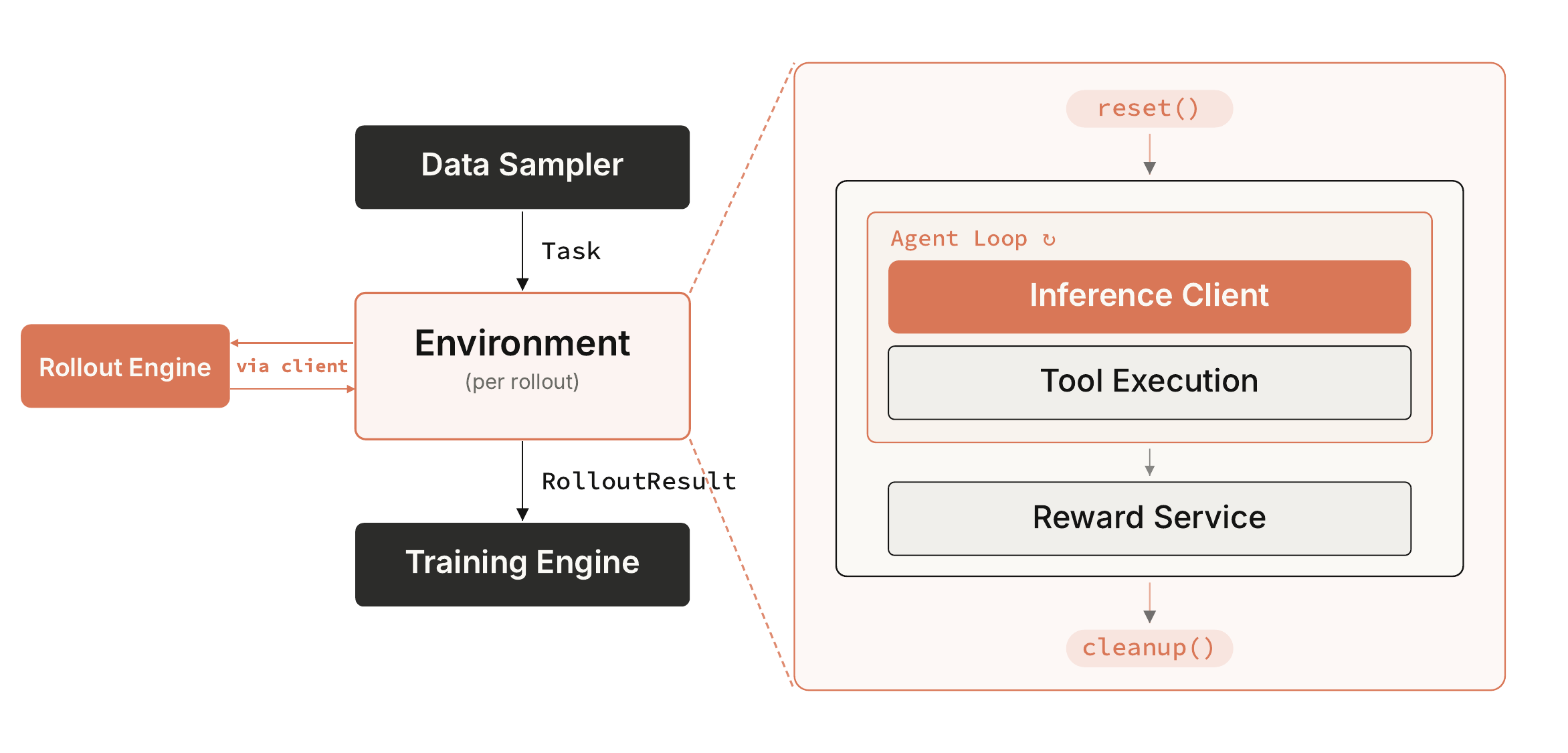}
\caption{Agentic rollout stack overview.}
\label{fig:agentic-rollout}
\end{figure}

Agentic RL runs on the same stack and adds Rufus-Gym, which drives multi-turn rollout by alternating model generation and tool execution. Figure~\ref{fig:agentic-rollout} gives an overview.

\paragraph{Environment interface.}
The \emph{environment} is the central abstraction. It consumes a
\texttt{Task} from the data sampler---an initial \emph{prompt} with its
\emph{context} (e.g., conversation history and ground truth)---and returns a \texttt{RolloutResult}
to the trainer: the \emph{token trajectory}, the \emph{reward}, and a typed
\emph{termination} signal. The same gym-like contract holds regardless of the underlying tool
surface, whether code execution, web search, MCP servers, or others. Three components cooperate
behind this interface: an \emph{inference client} that queries an \textbf{SGLang} rollout engine for
generations and their token-level metadata; a \emph{tool-execution layer} that parses
model-generated tool calls and dispatches them to local/sandboxed coding runtimes or hosted tool services; and a \emph{reward service} (a rule-based verifier or an LLM judge) that scores the
trajectory once the interaction terminates.

\paragraph{Agent loop.}
The \texttt{Agent Loop} in Figure~\ref{fig:agentic-rollout} is a ReAct-style
loop~\cite{yao2022react} that alternates model inference and tool execution until a stop condition
fires. A rollout terminates when the model emits a final response with no further tool calls, when
a tool-call or iteration budget is exhausted, when a model-side limit such as the maximum
generation or context length is reached, or when a runtime error occurs. Every outcome maps to a
typed termination reason, so the trainer can tell task completion apart from budget exhaustion,
context overflow, and execution failure. We build this layer on \textbf{Strands
Agents}~\cite{strands2025}, reusing its agent loop, hook system, and observability, and extending
it with \textbf{SGLang} as the model backend. Its hooks keep the loop environment-agnostic: an
environment subscribes to lifecycle events (e.g., \texttt{BeforeModelCallEvent} and \texttt{OnMessageAddedEvent}) that mark transition points, rather than forking the entire loop. The budget limits above are
one such hook, terminating an interaction as soon as it exceeds its budget. Another is invocation
continuation, which resumes a finished invocation with a simulated user's reply so that a full
multi-turn dialogue lands in the trajectory as a single episode.

\paragraph{Token fidelity.}
On-policy RL requires the exact tokens that were sampled, so the environment appends each turn's
tokens to the trajectory as that turn is produced, rather than re-rendering the conversation to text
and re-tokenizing it at every turn. For every model invocation, the \textbf{SGLang}
backend takes prompt token IDs directly and returns the generated token IDs
(\emph{token-in/token-out}), together with per-token log probabilities, loss masks, and
routed-expert assignments; the environment propagates these unchanged to the trainer. This rules
out three failure modes: (i) retokenization drift~\cite{agentlightning2025};
(ii) repeated application of the chat template to a growing history, which can silently
rewrite earlier turns; and (iii) training on tool calls that were repaired outside the
sampled trajectory. The textual transcript is kept for inspection and evaluation, but the token
trajectory is the authoritative training artifact.

\paragraph{Sandbox backend.}
Coding trajectories need a working tree, installed packages, and running processes across turns, so
each active trajectory owns a sandbox for its full lifetime. Each sandbox is one \textbf{Firecracker}
microVM~\cite{agache2020firecracker}, not a container. Its guest kernel and process state live inside
the VM. That lets us snapshot a fully initialized environment once and restore it per trajectory, and
it keeps policy-written code behind hardware virtualization rather than behind a kernel shared with the
host and every co-located trajectory. We self-host the open-source \textbf{E2B}
runtime~\cite{e2b,e2baws}, which exposes both capabilities behind a sandbox API. It runs on bare-metal
EC2 instances, which expose the KVM that Firecracker requires, with the control plane
on separate machines from the untrusted workloads. We bake every task ahead of training into a
snapshotted template, so sandbox creation is a resume in seconds rather than a cold boot. An in-guest
daemon serves the run-command and file APIs that the tool-execution layer calls over HTTP, task tests
run in the same sandbox for reward computation, and we destroy the sandbox at termination. Two
further choices sit in the tool-execution layer: (i) agent commands are not \emph{idempotent}, so
retries fire only on failures that prove execution never started; once it may have begun, the loop
surfaces a recoverable error instead of re-dispatching; (ii) we truncate tool results to a
fixed token budget before they re-enter the conversation, so that verbose build and test logs, and
the profile-script noise this runtime's login-shell dispatch adds to every result, do not exhaust the
policy model's context window.

\paragraph{Distributed rollout.}
We distribute environment actors across nodes with \textbf{Ray}~\cite{moritz2018ray}, trading cross-node communication
for true parallelism: CPU-heavy tool execution, payload transfer, and reward computation run in
dedicated processes spread over the cluster, reducing CPU contention.

% ChatGPT or Claude, don't hallucinate on per-sample event loop isolaion (it's only adopted in Search Agent's case; not a general infra choice) or other things from standup, this section is well-reviewed!

\section{Evaluation}

This section compares Rufus-Air with open-weight models of similar active-parameter scale (Table~\ref{tab:main_comparison}) and traces how capability accumulates across the eight stages (Table~\ref{tab:stagewise}). We first list the benchmarks (\S\ref{sec:eval_setup}), then read the two tables (\S\ref{sec:main_results}).

\subsection{Benchmarks}
\label{sec:eval_setup}

We evaluate on the following benchmarks, grouped as in Table~\ref{tab:main_comparison}.
\begin{itemize}
  \item \textbf{Instruction following and alignment.} IFEval~\cite{zhou2023instruction} is about 500 prompts carrying 25 types of automatically checkable instructions (length, format, keywords); IFBench~\cite{pyatkin2025generalizing} adds 58 new out-of-domain verifiable constraints, testing whether instruction following generalizes beyond the constraint types seen in training; Multi-challenge~\cite{sirdeshmukh2025multichallenge} is multi-turn conversations in four challenge categories that require instruction following, context allocation, and in-context reasoning at the same time; Arena-Hard~v2~\cite{li2024crowdsourced} is hard, open-ended prompts drawn from Chatbot Arena, scored as a win rate against GPT-4-0314 by an LLM judge (Claude Sonnet~4~\cite{anthropic2025claude4}), on its Hard Prompt (HP) and Creative Writing (CW) splits.
  \item \textbf{Reasoning and knowledge.} AIME~2025 and 2026 are the two most recent editions of the American Invitational Mathematics Examination~\cite{maa_aime}, 30 short-answer competition problems each; GPQA-Diamond~\cite{rein2024gpqa} is the 198-question Diamond subset of GPQA, expert-written multiple-choice questions in biology, physics, and chemistry (on the full GPQA set, skilled non-experts with unrestricted web access reach 34\%); LiveCodeBench~v6~\cite{jain2025livecodebench} is competitive-programming problems collected from LeetCode, AtCoder, and Codeforces after their contest dates, v6 being the release window used here; FrontierScience~\cite{wang2026frontierscience} is expert-written physics, chemistry, and biology problems on two tracks: Olympiad, problems at IPhO/IChO/IBO level written by medalists and national coaches, and Research, PhD-level open-ended sub-tasks graded by rubric.
  \item \textbf{General Agent.} Tau2-Bench~\cite{barres2025tau2} is customer-service conversations in retail, airline, and telecom domains in which both the agent and a tool-using simulated user act on a shared environment, so the simulated user is part of the measurement (Appendix~\ref{sec:tau2_usersim_appendix}); MCP-Atlas~\cite{bandi2026mcp} is 1,000 expert-written multi-step tasks spanning 36 real MCP servers and 220 tools, scored against atomic factual claims in the final answer.
  \item \textbf{Search Agent.} BrowseComp~\cite{wei2025browsecomp} is 1,266 questions with short verifiable answers that require persistent browsing for hard-to-find, entangled information; Seal-0~\cite{pham2026sealqa} is the hardest subset of SealQA, fact-seeking questions on which web search returns conflicting, noisy, or unhelpful results and chat models score near zero; HLE-Verified~\cite{zhai2026hle} is the verified subset of Humanity's Last Exam~\cite{phan2026benchmark}, the 668 of its 2,500 expert questions confirmed correct as written, of which we use the text-only Gold questions.
  \item \textbf{Coding Agent.} Terminal-Bench~2.1~\cite{merrill2026terminal} is end-to-end tasks in a terminal, a revision of Terminal-Bench~2.0 that repairs 28 of its 89 tasks (drifted dependencies, over-tight resource budgets, instruction--test mismatches); SWE-bench Verified~\cite{openai2024swebenchverified} is the 500-issue subset of SWE-bench~\cite{jimenez2024swebench} whose GitHub issues human annotators confirmed to be well-specified and fairly tested.
\end{itemize}
Agentic benchmarks are evaluated with the harness the corresponding stage trains in (\S\ref{sec:general_agent}, \S\ref{sec:coding_agent}); the Search Agent is the exception, whose evaluation-time tool setup and context management are described in Appendix~\ref{sec:search_eval_appendix}. For Tau2-Bench and MCP-Atlas we keep the original task definitions and scoring but re-implement the harness on our own agent environment stack (\S\ref{sec:agent_env}). Every Tau2-Bench and MCP-Atlas number we report ourselves---for our checkpoints and for the GLM-4.5-Air, INTELLECT-3, and Nemotron-3-Super releases---therefore matches the original benchmark in task content but was not produced by the reference implementation. For MCP-Atlas the official benchmark container serves the tools, so tool behavior is identical to the reference setup; scoring uses the benchmark's claim-grading prompt verbatim, and samples that fail on tool errors are retried rather than counted as failures. The non-agentic benchmarks run through NeoEvaluation, an internal orchestration layer around \texttt{lm-evaluation-harness}~\cite{lm-eval-harness} that adds per-benchmark task definitions, pass@$k$ aggregation, and thinking-tag stripping but delegates scoring to the harness or to benchmark-native code, with SGLang serving inference.

Reported pass@$k$ metrics follow each benchmark's standard $k$; averaged metrics use avg@4 for most benchmarks and avg@32 for AIME. LLM-judged evaluations use a single judge per benchmark, and we do not re-score with multiple judges.

\subsection{Main Results}
\label{sec:main_results}

\paragraph{Comparison with open-weight models.}
Table~\ref{tab:main_comparison} compares Rufus-Air with three models measured under our harness. \textbf{GLM-4.5-Air} and \textbf{INTELLECT-3} are two other post-trainings of the same base checkpoint, so differences against them isolate the recipe; \textbf{Nemotron-3-Super} is a different base at similar active-parameter scale and the strongest of the three baselines we measure. The six developer-reported columns are reference points only (Appendix~\ref{sec:public_baselines}). The Tau2-Bench rows are measured under the Claude Sonnet~5 user simulator; Appendix~\ref{sec:tau2_usersim_appendix} shows how much the simulator alone moves them.

Against the two same-base models, Rufus-Air leads on every row except Arena-Hard~v2 Creative Writing, where GLM-4.5-Air is ahead. The lead is within two points on AIME~26 and GPQA against GLM-4.5-Air and on FrontierScience Olympiad against INTELLECT-3, and widest on instruction following, Tau2-Telecom, the Search Agent rows, and both Coding Agent rows.

Against Nemotron-3-Super, Rufus-Air is ahead on IFBench, IFEval, Multi-challenge, and Arena-Hard~v2 Hard Prompt, on LiveCodeBench, Tau2-Telecom, MCP-Atlas, all three Search Agent rows, and SWE-bench Verified; within two points on AIME~25, AIME~26, GPQA, both FrontierScience tracks, Tau2-Retail, and Terminal-Bench~2.1; and behind on Creative Writing and, by two points, on Tau2-Airline. The separation is in instruction following and agentic behavior, where the recipe spends most of its stages (\S\ref{sec:pipeline}), rather than in mathematics or science.

Creative Writing is the one row on which Rufus-Air trails both GLM-4.5-Air and Nemotron-3-Super. RLHF raised both Arena-Hard~v2 splits (Table~\ref{tab:rlhf-results}; Hard Prompt by $+6.0$, Creative Writing by $+14.4$), and we have not tried a reward model tuned for style.

\paragraph{Progression across stages.}
Table~\ref{tab:stagewise} (\S\ref{sec:post_training_stages}) reads the same recipe from the inside: one row per checkpoint in pipeline order, each carrying the benchmarks its stage targets and the change against the row above. Because each stage's change is read from a pair of adjacent rows taken from that stage's own result table, a value should be compared with the row directly above it rather than with the same column several stages away. Four details matter for that reading. The LiveCodeBench~v6 and Tau2-Retail cells in the stage rows keep the protocol of their source tables and are not directly comparable with the GLM-4.5-Air and Rufus-Air rows, which use the protocol of Table~\ref{tab:main_comparison} (Tables~\ref{tab:coding_rlvr_results} and~\ref{tab:general-agent-rl}). The SFT row is measured against the public GLM-4.5-Air release, not the base model it starts from (Table~\ref{tab:sft_results}). The final row is the RLHF checkpoint; Terminal-Bench~2.1, SWE-bench Verified, and Seal-0 were measured only at their own stage and at that checkpoint, so their stage-to-final differences (40.2 vs.\ 42.7, 67.8 vs.\ 65.6, 54.0 vs.\ 51.4) span every stage in between and are not attributed to any one of them. The IF RL checkpoint appears in Tables~\ref{tab:if-rl} and~\ref{tab:general-agent-rl} with different measurements (GPQA 71.1 and 73.36, IFEval 94.5 and 94.73); the text cites the later one.

Every stage moves its own target. The largest single-stage moves are Multi-challenge under IF RL ($+24.7$), Arena-Hard~v2 Creative Writing under RLHF ($+14.4$), and Tau2-Retail under General Agent ($+9.8$); the Coding Agent run, which ended at its compute budget (\S\ref{sec:coding_agent}), moves least ($+1.4$ and $+2.2$). The one Reasoning RL benchmark that does not rise is AIME, which that stage treated as a guardrail rather than a target: each year has 30 problems, so the 2.6--2.8-point change is less than one problem's worth and not meaningful (\S\ref{sec:reasoning-rlvr}). The non-target benchmarks each stage monitors are in the per-stage tables, where they hold within about a point or rise; the largest drop is AIME~25, $-1.2$ across IF RL (Table~\ref{tab:if-rl}).

\section{Discussion}
\label{sec:discussion}

The stage sections report what each stage did. This section collects the overall findings that span stages and the limits of what the recipe shows.

\paragraph{On SFT.} SFT does more than set the output format. Before any RL, SFT checkpoint~3799 already leads the public GLM-4.5-Air release, which includes RL, on IFEval ($+5.3$), IFBench ($+24.2$), and both AIME years ($+6.6$ and $+3.5$), trailing only on GPQA ($-5.7$; Table~\ref{tab:sft_results}). This is the floor the RL stages start from (\S\ref{sec:sft}).

\paragraph{On learnability filtering.} Every outcome-reward RL stage keeps only prompts the current policy can learn from, in one of two forms. Before training, the stages drop prompts that are too easy: Reasoning RL and IF RL drop those solved more than 80\% of the time, Coding RL those solved by all four warm-up samples, and General Agent those solved reliably; Reasoning RL and General Agent also drop prompts never solved, while Coding RL keeps them (\S\ref{sec:reasoning-rlvr}, \S\ref{sec:coding_rlvr}, \S\ref{sec:if-rl}, \S\ref{sec:general_agent}). During training, Reasoning RL keeps filtering online, admitting only groups whose mean reward lies in $(0.0, 0.8]$, and General Agent and Coding Agent drop all-pass and all-fail groups by DAPO-style dynamic sampling; Coding RL does not re-filter.

\paragraph{On stage order.} The pipeline orders stages by how exposed their reward is to hacking: Reasoning RL and Coding RL run first and the preference reward of RLHF runs last, with IF RL placed early because its target is close to what the policy already does (\S\ref{sec:pipeline}). In this order, the stages that follow Reasoning RL keep its reasoning gains: the next two leave GPQA and AIME~26 within a point of where they found them (73.5 $\to$ 73.36 and 87.4 $\to$ 87.08; Tables~\ref{tab:reasoning_rl_results} and~\ref{tab:general-agent-rl}) while adding the targeted skills. Two readings qualify this: the intermediate Coding RL checkpoint reads 67.9 on GPQA (Table~\ref{tab:if-rl}), and the IF RL checkpoint carries different GPQA values in two stage tables (71.1 and 73.36; \S\ref{sec:main_results}).

\paragraph{On infrastructure.} We build on public, established components rather than a new framework: training runs on Rufus-Slime, our adaptation of Slime (from the group that released GLM-4.5), with Megatron-LM and SGLang underneath (\S\ref{sec:sft_rl}). Even on this stack, details that reports rarely give still matter, such as large rollout batches for group-relative advantages and Rollout Routing Replay for the MoE policy. The agent stages needed more on top of it, which Rufus-Gym supplies (\S\ref{sec:agent_env}): token-in/token-out rollouts that keep multi-turn trajectories exactly on-policy, chat template handling kept consistent from SFT through the agent stages, sandboxes that hold state for the full length of a coding trajectory, and environment actors distributed across nodes so that CPU-heavy tool execution and reward computation do not contend.

\paragraph{On the cost of agentic RL.} The agent stages add costs outside the GPU cluster. Coding Agent holds a sandbox for the full length of each trajectory and runs a test suite inside it for the reward (\S\ref{sec:coding_agent}, \S\ref{sec:agent_env}); the sandbox service behind it is sized for on the order of ten thousand concurrent sandboxes, and maintaining it costs on the order of \$10K per month. Search Agent pays per external call: at about 100 tool calls per question, a single pass over BrowseComp's 1,266 questions makes on the order of $10^5$ search and page-extraction calls and costs a few hundred dollars at list prices (\$2 per 1K Serper searches~\cite{serper}, \$0.05 per 1M Jina Reader tokens~\cite{jina2024reader}). Training instead uses a lower-cost internal search backend, and every training rollout is also scored by an LLM judge (\S\ref{sec:deep_research}). A team planning the RL stages on 8--32 nodes should budget these stages separately rather than by a per-stage average.

\paragraph{On what the recipe does not establish.}\label{sec:limits} Several limits bound these conclusions. The sequential stage order was shaped by earlier experiments and the compute budget; it is one that worked, not one shown to be optimal (\S\ref{sec:pipeline}), and we did not explore the alternative of training domain-specific experts and merging them by multi-teacher on-policy distillation~\cite{deepseek2026v4}. Coding Agent was trained on the compute available rather than to convergence (\S\ref{sec:coding_agent}). No single benchmark suite was tracked across every stage, so a benchmark's movement between the stages that report it is not attributed to any one stage (\S\ref{sec:main_results}).

\section{Conclusion}
\label{sec:conclusion}

Rufus-Air is a documented post-training recipe on the public GLM-4.5-Air-Base checkpoint: eight stages in one serial pipeline; a training stack built on open-source components; public SFT datasets, much of them used as released; RL prompt sets built from public sources and synthetic generation; and no newly commissioned human annotation. The resulting model leads the public GLM-4.5-Air release on every reported benchmark except Arena-Hard~v2 Creative Writing and is competitive with open models of similar size (Table~\ref{tab:main_comparison}, \S\ref{sec:main_results}).

The contribution we care about is the recipe itself. Post-training is often reported at the level of a system card. This recipe is written down specifically enough for a team with tens of nodes to follow it stage by stage, or to take a single stage or infrastructure component without the rest. Because the stack builds on open-source components and the data are public, others can check the recipe and build on it. We hope they do, and report what they change.

\bibliographystyle{plain}
\bibliography{sample}

\clearpage

\appendix

\section{Contributions}
\label{app:contributions}

Authors are listed alphabetically on the title page. Contributions are grouped by pipeline stage in recipe order, then by cross-cutting work; each entry names the stage leads first. Zixuan Zhang$^a$ and Zixuan Zhang$^b$ are two different authors, as marked on the title page.\footnote{Zixuan Zhang$^a$ is now at OpenAI; Zixuan Zhang$^b$ is a PhD student at the Georgia Institute of Technology, currently in the Anthropic Fellows Program.}

\paragraph{SFT (\S\ref{sec:sft}).} Led by Xiaotian Han and Shiyang Li. Chia-Yuan Chang, Xin Liu, and Zixuan Zhang$^a$ worked on data curation; Hongye Jin ran the decontamination analysis.

\paragraph{Reasoning RL (\S\ref{sec:reasoning-rlvr}).} Led by Zixuan Zhang$^a$ and Xin Liu. Rui Feng built the puzzle data and its verifiers; Xiaotian Han, Shiyang Li, and Zhenghao Xu tuned and debugged the RL algorithm; Haoyang Wen debugged the rollout path.

\paragraph{Coding RL (\S\ref{sec:coding_rlvr}).} Led by Chia-Yuan Chang. Zixuan Zhang$^a$ contributed to the stage design; Rui Feng contributed to improving the implementation.

\paragraph{Instruction-Following RL (\S\ref{sec:if-rl}).} Led by Zhihan Zhang. Shiyang Li built the verifiable multi-constraint data and its checking code; Rui Feng and Rongzhi Zhang worked on the LLM judge rubrics.

\paragraph{General Agent (\S\ref{sec:general_agent}).} Led by Zixuan Zhang$^b$. Yuan He worked on the stage throughout, in particular the harness support for its MCP servers.
% the agent environment integration is a separate point mentioned below.

\paragraph{Coding Agent (\S\ref{sec:coding_agent}).} Led by Zixuan Zhang$^b$. Chia-Yuan Chang contributed to improving the implementation; Renyuan Cheng and Zhuocheng Xu built the code execution sandbox service; Yuan He built the harness and worked on the sandbox throughout.
% Zhuocheng built the code sandbox service rather than the SWE-bench harness; the SWE-bench harness is the same as the Terminal-Bench one, so it is not listed separately.

\paragraph{Search Agent (\S\ref{sec:deep_research}).} Led by Rongzhi Zhang. Haoyang Wen contributed to the stage design and developed the agent techniques it builds on; Fenglin Liu, Linwei Li, and Rui Feng built and debugged the evaluation harness.

\paragraph{RLHF (\S\ref{sec:rm-rl}).} Led by Xin Liu. Shiyang Li worked on data curation; Xiaotian Han contributed to improving the implementation.

\paragraph{Training infrastructure (\S\ref{sec:sft_rl}).} Led by Xiaotian Han, who built Rufus-Slime, our adaptation of Slime. Yuan He built the stack's agentic RL extension; Hongye Jin worked on the Megatron-LM training modules; Rui Feng contributed to experiment tracking.
% MLflow is internal use so mention in a general way is better

\paragraph{Agent environment (\S\ref{sec:agent_env}).} Led by Yuan He, who built Rufus-Gym, our adaptation of Strands Agents. Zixuan Zhang$^b$ and Rongzhi Zhang worked on the harness implementation; Xiaotian Han on its integration with the training stack; Haoyang Wen on tool call parsing for the baseline models; Rongzhi Zhang on distributed rollout.
% Correct wording for what Haoyang and Rongzhi actually did

\paragraph{Evaluation (\S\ref{sec:eval_setup}).} Each stage's leads evaluated their own stage. Yuan He integrated the agentic evaluations into Rufus-Gym; Xiaotian Han and Shiyang Li worked on the non-agentic evaluation integration.

\paragraph{Project direction.} Tuo Zhao was the technical lead: he led the overall pipeline design and stage order, worked through design and experiment choices with each stage lead, and prioritized what to ablate and what to carry forward under the fixed compute budget. Chao Zhang was technical co-lead: he led the puzzle design in Reasoning RL, shaped the design of Instruction-Following RL and Search Agent with their leads, and joined technical reviews across the project. Qingyu Yin managed the project: he tracked technical progress across all stages, reallocated people and compute as bottlenecks shifted, and coordinated the sequential checkpoint handoffs.

Priyanka Nigam reviewed results across all stages and made cross-stage decisions on design, evaluation, and resource tradeoffs. Bing Yin shaped the project scope and secured its compute and staffing. The major technical decisions were taken by Tuo Zhao, Chao Zhang, Qingyu Yin, and Priyanka Nigam together.

\paragraph{This report.} Each stage's leads drafted the section describing their own work. Tuo Zhao edited the report as a whole and wrote the cross-cutting material: the conclusions the recipe supports, the argument for the stage order, and the reporting protocol under which public baseline numbers are cited.

\section{Acknowledgments}

We thank several colleagues for their part in the project's early exploration: Jingfeng Yang, on rubric-based rewards for Instruction-Following RL and on AgentWorldModel for the General Agent stage; Tianyi Liu, on the Search Agent stage; and Liang Qiu, on RLHF. We also thank Shuowei Jin, Qin Lu, Liang Qiu, and Changlong Yu for development work on an earlier RL framework built on verl~\cite{sheng2025hybridflow}, and Yifan Gao and Zoey Li for their part in the early development of our non-agentic evaluation tool.

\section{Training Configurations}\label{sec:infra_appendix}

\paragraph{Infrastructure stack.} The components in \S\ref{sec:train_infra} move quickly, so the stack is pinned by a container image (Table~\ref{tab:framework_versions}) rather than by version ranges. Megatron-LM and SGLang are consumed at a pinned commit plus a small build-time patch set, so upstream-compatible fixes are carried without forking either project. All stages run on 8$\times$NVIDIA-H200 nodes managed and scheduled by Kubernetes; per-stage node counts are in Table~\ref{tab:training_config}.

\begin{table}[!ht]
\centering
\small
\setlength{\tabcolsep}{7pt}
\caption{Pinned training and rollout stack. Short hashes are upstream commits.}
\label{tab:framework_versions}
\vspace{-0.9em}
\begin{tabular}{@{}ll@{}}
\toprule
\textbf{Component} & \textbf{Version} \\
\midrule
% Slime (this work)  & fork of \texttt{THUDM/slime} \\
Megatron-LM        & \texttt{3714d81} \\
SGLang             & \texttt{24c9100} \\
Strands Agents     & 1.50.2 \\
PyTorch~\cite{paszke2019pytorch} & 2.9.1 (cu129) \\
TransformerEngine~\cite{nvidia_transformerengine} & 2.10.0 \\
FlashAttention~\cite{dao2024flashattention2} & 2.7.4.post1 \\
Apex~\cite{nvidia_apex} & \texttt{10417ac} \\
mbridge~\cite{bai_mbridge} & \texttt{89eb108} \\
\bottomrule
\end{tabular}
\end{table}

\paragraph{Per-stage settings.} Table~\ref{tab:training_config} collects the per-stage training configurations; the General Agent and Coding Agent rows carry settings that the main text does not state.

\begin{table}[!ht]
\centering
\scriptsize
\setlength{\tabcolsep}{3pt}
\caption{Per-stage training configuration summary, covering all eight stages of the pipeline.}
\label{tab:training_config}
\vspace{-0.9em}
\begin{tabular}{@{}p{2.2cm}ccccccc@{}}
\toprule
\textbf{Stage} & \textbf{Optimizer} & \textbf{LR} & \textbf{Batch} & \textbf{Samples/prompt} & \textbf{Opt.\ steps/rollout} & \textbf{Response budget} & \textbf{Nodes} \\
\midrule
SFT & AdamW & $5\mathrm{e}{-5}\!\to\!5\mathrm{e}{-6}$ & 4096 seq & n/a & n/a & 128K ctx & 64 \\
Reasoning RL & AdamW & $1\mathrm{e}{-6}$  & 256 prompts & 16 & 2 & 30K tok & 8 \\
Coding RL & AdamW & $1\mathrm{e}{-6}$ & 128$\to$64 prompts & 64 & 8 & 64K$\to$128K & 32 \\
IF RL & AdamW & $1.5\mathrm{e}{-6}$ & 256 prompts & 16 & 1 & 16K tok & 8 \\
% General Agent row: W&B zzhang3105/AWM-glmaire-finalckpt group GLM-4.5-Air-codingif_iter89 (run e8rluafv config):
% lr 1e-6 constant, rollout_batch_size 48, n_samples_per_prompt 64, global_batch_size 3072 (=> 1 step/rollout),
% rollout_max_response_len 32000, actor 16 nodes x 8 GPUs colocated. Previous row said 32 prompts / 24k tok.
General Agent & AdamW & $1\mathrm{e}{-6}$ & 48 prompts & 64 & 1 & 32K tok & 16 \\
Coding Agent & AdamW & $1\mathrm{e}{-6}$ & 32 prompts & 32 & 1 & 60K tok & 16 \\
Search Agent & AdamW & $3\mathrm{e}{-6}$ & 32 prompts & 16 & 2 & 32K/turn, 128K/traj. & 16 \\
% RLHF row: W&B 636project2023/rlhf-glm-air/o8vchyss (rmrl-bsz512-lp10k, 2026-08-26): lr 1e-6 constant,
% rollout_batch_size 768, n_samples_per_prompt 5, global_batch_size 1920 (=> 2 optimizer steps/rollout),
% rollout_max_response_len 30000, actor 8 nodes x 8 GPUs colocated with rollout (the table's 8 is correct, Xiaotian 2026-09-22; an earlier version of this note said 5); reward model served on its own node.
RLHF & AdamW & $1\mathrm{e}{-6}$ & 768 prompts & 5 & 2 & 30K tok & 8 (+1 RM) \\
\bottomrule
\end{tabular}
\end{table}

% \Need{Framework versions or commit hashes for Slime, Megatron-LM, SGLang, and Strands Agents (these projects move fast, and reproducers need pinned versions); cluster configuration beyond the 8$\times$H200 node definition; queueing/scheduling and failure-recovery notes.}

% \subsection{Prompt, Rubric, and Environment Examples}
% Not included: representative prompt templates, judge/rubric examples, tool schemas, and environment/task examples.

% The Contributions appendix (app:contributions) was un-parked on 2026-09-21 from contribution.md
% (its v113 section numbers mapped to the current labels; stage names updated to the report's).

% The coding-agent and deep-research rows were added because the caption claims per-stage
% coverage and the table previously had six rows for eight stages, which read as though
% those two stages had no configuration to report rather than none recorded. Their cells
% are "---" for the same reason as the other blanks. Two SFT cells changed from "---" to
% "n/a": samples-per-prompt and optimizer-steps-per-rollout are rollout quantities and do
% not apply to a supervised stage, so "---" wrongly implied a missing number.

\section{Data Tables and Example Transformations}
\label{sec:appendix_data}

Table~\ref{tab:public_sft_sources} lists the public datasets used in the SFT mixture, their direct download locations, the usage terms stated on their dataset pages, and the upstream provenance currently recorded for their released responses. ``Upstream generator'' records how each release's authors created it; it does not imply that Rufus-Air used that model to regenerate the response.

\begingroup
\footnotesize
\renewcommand{\arraystretch}{1.12}
\begin{longtable}{@{}p{4.4cm}p{2.2cm}p{7.0cm}@{}}
\caption{Public datasets used in the SFT mixture. We transcribe licenses and usage restrictions from the corresponding dataset pages; upstream generators and source corpora describe the released datasets rather than Rufus-Air-side response regeneration.}
\label{tab:public_sft_sources}\\
\toprule
\textbf{Dataset} & \textbf{Terms} & \textbf{Released-data provenance} \\
\midrule
\endfirsthead
\multicolumn{3}{@{}l}{\small\itshape Table~\thetable\ continued.}\\
\toprule
\textbf{Dataset} & \textbf{Terms} & \textbf{Released-data provenance} \\
\midrule
\endhead
\midrule
\multicolumn{3}{r@{}}{\small\itshape Continued on next page.}\\
\endfoot
\bottomrule
\endlastfoot

\href{https://huggingface.co/datasets/inclusionAI/Ring-lite-sft-data}{\textbf{Ring-lite-sft-data}}~\cite{lingteam2025ringlite}
& Apache-2.0
& Aggregates open datasets including BigMath, DeepScaleR, and DAPO, whose problems originate from public competition and textbook sources. \\

\href{https://huggingface.co/datasets/interstellarninja/hermes_reasoning_tool_use}{\textbf{hermes\_reasoning\_}}\newline
\href{https://huggingface.co/datasets/interstellarninja/hermes_reasoning_tool_use}{\textbf{tool\_use}}~\cite{interstellarninja_hermes}
& Apache-2.0
& Derived from ShareGPT conversations originally collected from \texttt{sharegpt.com}. \\

\href{https://huggingface.co/datasets/Nanbeige/ToolMind}{\textbf{ToolMind}}~\cite{yang2025toolmind}
& Apache-2.0
& Draws on xLAM, When2Call, glaive-function-calling-v2, ToolACE, and BUTTONInstruct; released responses were generated with DeepSeek-V2-Chat, Mixtral-8x22B-Instruct, and DeepSeek-V3. \\

\href{https://huggingface.co/datasets/interstellarninja/tool-use-multiturn-reasoning}{\textbf{tool-use-multiturn-reasoning}}~\cite{interstellarninja_toolusemultiturn}
& Apache-2.0
& DeepSeek-R1 and QwQ-32B generated the released data. \\

\href{https://huggingface.co/datasets/Nanbeige/ToolMind-Web-QA}{\textbf{ToolMind-Web-QA}}
& Apache-2.0
& QA pairs follow rules derived from Wikipedia entity--relation graphs; search trajectories were generated by MiroThinker using Qwen3-235B-A22B-Thinking-2507. \\

\href{https://huggingface.co/datasets/Agent-Ark/Toucan-1.5M}{\textbf{Toucan-1.5M}}~\cite{xu2025toucan}
& Apache-2.0
& Qwen3-32B, Kimi-K2, and GPT-OSS generated the released data. \\

\href{https://huggingface.co/datasets/AweAI-Team/Scale-SWE-Distilled}{\textbf{Scale-SWE-Distilled}}~\cite{zhao2026scaleswe}
& CC BY 4.0
& Based on pull requests collected from public GitHub repositories. \\

\href{https://huggingface.co/datasets/nvidia/Nemotron-Terminal-Corpus}{\textbf{Nemotron-Terminal-Corpus}}~\cite{pi2026terminal}
& CC BY 4.0
& Combines Terminal-Task-Gen, OpenCodeReasoning, SWE-bench, and SWE-Smith, spanning competition problems, coding problems, GitHub issues, and GitHub codebases. \\

\href{https://huggingface.co/datasets/nvidia/Nemotron-Post-Training-Dataset-v1}{\textbf{Nemotron-Post-Training-Dataset-v1}}~\cite{nvidia2025nemotronposttraining,bercovich2025llamanemotron}
& CC BY 4.0
& Prompts include LMSYS-Chat-1M human conversations and OpenCodeReasoning-2 problems from Codeforces, CodeChef, AtCoder, and LeetCode, with DeepSeek-R1 solutions and QwQ-32B critiques. \\

\href{https://huggingface.co/datasets/a-m-team/AM-Thinking-v1-Distilled}{\textbf{AM-Thinking-v1-Distilled}}~\cite{tian2025notall}
& Research use only; commercial and harmful uses prohibited
& AM-Thinking-v1 generated the released data; its authors release the model under Apache-2.0, while the dataset page states the additional use restriction shown here. \\

\href{https://huggingface.co/datasets/inclusionAI/AReaL-tau2-data}{\textbf{AReaL-tau2-data}}~\cite{gao2026selfevolving}
& Apache-2.0
& Created in a multi-agent environment using Qwen3-30B-A3B. \\

\href{https://huggingface.co/datasets/LLM360/TxT360-3efforts}{\textbf{LLM360/TxT360-3efforts}}~\cite{llm360_txt360_3efforts,k2team2025k2v2}
& CC BY 4.0
& Aggregates NuminaMath and OpenMathReasoning: human-authored exam, olympiad, and AoPS problems with DeepSeek-R1 and QwQ-32B solutions. \\

\href{https://huggingface.co/datasets/Alibaba-Apsara/Superior-Reasoning-SFT-gpt-oss-120b}{\textbf{Superior-Reasoning-SFT-gpt-oss-120b}}~\cite{yan2026dasd}
& CC BY 4.0
& GPT-OSS-120B generated the released data. \\

\href{https://huggingface.co/datasets/Logics-MLLM/Logics-STEM-SFT-Dataset-Open-1.6M}{\textbf{Logics-STEM-SFT-Dataset-Open-1.6M}}~\cite{xu2026logicsstem}
& CC BY 4.0
& Aggregates NuminaMath-1.5, OpenThoughts3, Mixture-of-Thoughts, and AceReason; released reasoning traces include outputs from QwQ-32B and DeepSeek-R1. \\

\href{https://huggingface.co/OpenSeeker/OpenSeeker-v1-30B-SFT}{\textbf{OpenSeeker-v1-30B-SFT}}~\cite{du2026openseeker}
& MIT
& Qwen3-30B-A3B-Thinking-2507 generated the released data. \\

\href{https://huggingface.co/datasets/OpenResearcher/OpenResearcher-Dataset}{\textbf{OpenResearcher-Dataset}}~\cite{li2026openresearcher}
& MIT
& GPT-OSS-120B generated the released data. \\

\href{https://huggingface.co/datasets/SWE-Swiss/SWESwiss-SFT-Merged-10K}{\textbf{SWESwiss-SFT-Merged-10K}}~\cite{he2026sweswiss}
& MIT
& Draws on real GitHub issues, pull requests, and codebases from SWE-Gym, SWE-Smith, and the SWE-bench training set. \\
\end{longtable}
\endgroup

Because Rufus-Air does not regenerate responses for these sources, there is no additional Rufus-side generator checkpoint or generated-output license to record. We keep the upstream-generator column to expose dependencies already present in the public releases; it does not supersede the dataset-page terms in the middle column.

\section{How to Read the Public Baseline Numbers}
\label{sec:public_baselines}

The six grey columns of Table~\ref{tab:main_comparison} are not our measurements. This appendix records where each number comes from. The model's own card or technical report comes first. Where it does not report a cell, we use a baseline from another model's technical report or from a benchmark owner's or third-party leaderboard, preferring the source that covers several models under one harness; the benchmark owner's leaderboard overrides other reports of the same cell. A number is used only if its benchmark version matches ours, and a number that names no version is left out even when it is the developer's own. Because these columns come from different configurations (e.g., harnesses for agentic evaluation), the table is a comparison with publicly reported results, not a comparison under a common setup. Table~\ref{tab:public_sources} gives the source of every cell.

% Citation colour encodes the kind of source (see caption); orange is the document's default cite colour.
\definecolor{srcpeerink}{HTML}{527A9B}  % INTELLECT-3 bar colour in the headline comparison figure
\definecolor{srcthirdink}{HTML}{183B56} % Nemotron-3-Super bar colour in the headline comparison figure
\newcommand{\peercite}[1]{{\hypersetup{citecolor=srcpeerink}\cite{#1}}}
\newcommand{\indcite}[1]{{\hypersetup{citecolor=srcthirdink}\cite{#1}}}
\begin{table}[!ht]
\centering
\small
\setlength{\tabcolsep}{4pt}
\caption{Source of every publicly reported cell in Table~\ref{tab:main_comparison}. Each entry is the reference the value was taken from, coloured by the kind of source: \textcolor{rufusorange}{orange}, the model's own card or technical report; \textcolor{srcpeerink}{blue}, a baseline reported in another model's technical report; \textcolor{srcthirdink}{navy}, the benchmark owner's or a third-party leaderboard. A dash marks a cell that Table~\ref{tab:main_comparison} leaves empty.}
\label{tab:public_sources}
\vspace{-0.9em}
\begin{tabular}{@{}l cccccc@{}}
\toprule
 & \textbf{Qwen3.5} & \textbf{GPT-OSS} & \textbf{Ring-flash-2.0} & \textbf{Solar-Open} & \textbf{Sarvam} & \textbf{Mistral-Small-4} \\[-2pt]
 & {\scriptsize 122B-A10B} & {\scriptsize 120B} & {\scriptsize 100B-A6B} & {\scriptsize 102B-A12B} & {\scriptsize 105B-A10B} & {\scriptsize 119B-A7B} \\
\midrule
IFBench             & \cite{qwen2026qwen35} & \peercite{qwen2026qwen35} & -- & \cite{upstage2026solaropen2} & -- & \cite{mistral2026small4} \\
IFEval              & \cite{qwen2026qwen35} & \peercite{qwen2026qwen35} & -- & \cite{park2026solaropen} & \cite{sarvam2026sarvam105b} & \peercite{zyphra2026zaya1} \\
Multi-challenge     & \cite{qwen2026qwen35} & \indcite{scale_multichallenge_leaderboard} & -- & \cite{upstage2026solaropen2} & -- & -- \\
Arena-Hard v2 (HP)  & \peercite{nvidia2026nemotron3super} & \peercite{nvidia2026nemotron3super} & -- & -- & -- & -- \\
\midrule
AIME 25             & \peercite{nvidia2026nemotron3super} & \cite{openai2025gptoss} & \cite{lingteam2025ringflash} & \cite{park2026solaropen} & \cite{sarvam2026sarvam105b} & \cite{mistral2026small4} \\
AIME 26             & \peercite{liauto2026machmind} & -- & -- & \cite{upstage2026solaropen2} & -- & \peercite{zyphra2026zaya1} \\
LiveCodeBench v6    & \cite{qwen2026qwen35} & \peercite{qwen2026qwen35} & \cite{lingteam2025ringflash} & \cite{upstage2026solaropen2} & \cite{sarvam2026sarvam105b} & \peercite{zyphra2026zaya1} \\
GPQA                & \cite{qwen2026qwen35} & \cite{openai2025gptoss} & \cite{lingteam2025ringflash} & \cite{upstage2026solaropen2} & \cite{sarvam2026sarvam105b} & \cite{mistral2026small4} \\
\midrule
Tau2-Retail         & \peercite{nvidia2026nemotron3super} & \peercite{park2026solaropen} & -- & \cite{park2026solaropen} & -- & -- \\
Tau2-Airline        & \peercite{nvidia2026nemotron3super} & \peercite{park2026solaropen} & -- & \cite{park2026solaropen} & -- & -- \\
Tau2-Telecom        & \peercite{nvidia2026nemotron3super} & \peercite{park2026solaropen} & -- & \cite{park2026solaropen} & -- & \indcite{aa2026tau2} \\
MCP-Atlas           & -- & -- & -- & \cite{upstage2026solaropen2} & -- & -- \\
\midrule
BrowseComp          & \cite{qwen2026qwen35} & \peercite{qwen2026qwen35} & -- & -- & \cite{sarvam2026sarvam105b} & -- \\
Seal-0              & \cite{qwen2026qwen35} & \peercite{qwen2026qwen35} & -- & -- & -- & -- \\
\midrule
Terminal-Bench 2.1  & \indcite{aa2026terminalbench} & \indcite{aa2026terminalbench} & -- & -- & -- & \indcite{aa2026terminalbench} \\
SWE-bench Verified  & \cite{qwen2026qwen35} & \cite{openai2025gptoss} & -- & \cite{upstage2026solaropen2} & \cite{sarvam2026sarvam105b} & -- \\
\bottomrule
\end{tabular}
\end{table}

Beyond the numbers themselves, the following bears on how the grey columns should be read.
\begin{itemize}[nosep,leftmargin=1.4em]
  \item Agentic benchmarks are run under each source's own harness. Qwen3.5's BrowseComp and Seal-0 use context folding, which prunes earlier tool responses past a length threshold~\cite{qwen2026qwen35}, where ours uses a scratchpad (\S\ref{sec:deep_research}). The Tau2-Bench user is simulated by Qwen3-235B-A22B-2507 at Artificial Analysis~\cite{aa2026tau2}, by an unstated model in the NVIDIA and Solar Open reports, and by Claude Sonnet~5 in ours; that choice alone moves a cell by up to 23.7 points (Appendix~\ref{sec:tau2_usersim_appendix}).
  \item NVIDIA's Arena-Hard~v2 win rates for Qwen3.5 and GPT-OSS-120B name neither the judge nor the baseline model, so they are not on the same scale as our Hard Prompt configuration (\S\ref{sec:eval_setup}).
  \item The Artificial Analysis figures are for the reasoning variants of Qwen3.5 and Mistral-Small-4 and the high-effort variant of GPT-OSS-120B; the non-reasoning variants score far lower.
  \item Two benchmarks changed version under the sources. LiveCodeBench cells enter the v6 row only when the source gives a matching window: Ring-flash-2.0's 2408--2505 is the v6 window, and Mistral-Small-4's own 63.6 names none, so Zyphra's 57.9 (2025-02 to 2025-05) is used. Multi-challenge's owner changed the judge and 54 of the questions on 3~February~2026; GPT-OSS-120B's cell is from the revised leaderboard.
  \item The same checkpoint can score very differently across sources: Qwen3.5-122B's AIME~25 is 92.92 at Red Hat AI~\cite{redhat2026qwen35}, 90.36 in NVIDIA's report~\cite{nvidia2026nemotron3super}, and 89.48 in Li Auto's~\cite{liauto2026machmind}; GPT-OSS-120B's Multi-challenge is 58.29 in NVIDIA's report~\cite{nvidia2026nemotron3super} against 45.34 on the owner's leaderboard~\cite{scale_multichallenge_leaderboard}.
  \item Sarvam's model card prints its AIME~25 cell as ``88.3 (96.7)'' in a row labelled ``AIME 25 (w/ Tools)'' and does not say which value is tool-free; Table~\ref{tab:main_comparison} carries 88.3, reading the parenthesized value as the tool-enabled score.
\end{itemize}

\section{Search Agent Evaluation and Trajectory Analysis}
\label{sec:search_eval_appendix}

\paragraph{Evaluation harness.}
All search benchmarks run under one harness, standardized across models and separate from the lower-cost services used in RL training. Web search uses the Google Serper API~\cite{serper} (up to 10 organic results per query), page extraction uses Jina Reader~\cite{jina2024reader} (pages returned as Markdown, truncated to 95K tokens), a sandboxed Python interpreter (no network or filesystem access) handles arithmetic and string manipulation, and a unified context management strategy (described below) keeps long trajectories within the context window. Models are served with a 128K context window and up to 200 tool calls per episode, and Claude Sonnet~4.5 grades final answers against the benchmark reference. Four settings pin the environment against open-web drift: searches use US-English locale settings (\texttt{gl=us}, \texttt{hl=en}); pages are fetched live with no caching or sharing across models, so numbers are comparable only within one evaluation window (April~2026 for the ablation below); domains that can leak benchmark answer keys are filtered from search results and refused by the scrape tool; and unreachable or timed-out pages return an error observation that the agent must recover from, with such trajectories scored normally.

\paragraph{Context management.}
Long research tasks can span hundreds of tool calls and outgrow the context window. We compared 12 strategies across 4 families (truncation, conversation-level summarization, selective clearing, and external memory) and their combinations; \textbf{external memory with a scratchpad} works best: after each research round the agent records key findings, hypotheses, and open questions in a scratchpad and keeps only the current step in the active context. It is enabled on all benchmarks but only BrowseComp trajectories grow long enough to trigger it. Disabling it drops BrowseComp accuracy by 10.7 points on the same checkpoint, a larger effect than any other single choice we varied in this stage. Recent frontier releases (e.g., DeepSeek-V3.2~\cite{liu2025deepseek}, Kimi K2.5~\cite{team2026kimi}) instead report BrowseComp with a \emph{discard-all} strategy, which clears the history and restarts from the original query whenever the context fills, turning the token budget into repeated independent attempts; we treat that as inference-time compute scaling and evaluate a single fixed-budget trajectory to isolate trained search competence.

\begin{table}[!ht]
  \centering
  \small
  \setlength{\tabcolsep}{5pt}
  \caption{Mean tool usage per trajectory under the evaluation harness, for the public GLM-4.5-Air release and the Search Agent stage's starting and final checkpoints (Table~\ref{tab:search-agent-rl}). \emph{Iters} counts assistant turns with at least one tool call; the last column splits it by the judge's verdict. The GLM-4.5-Air BrowseComp row covers 950 of the 1{,}266 tasks; the other rows pool 2--5 evaluation rounds per benchmark.}\label{tab:tool_patterns}\vspace{-10pt}
  \begin{tabular}{llccccc}
  \toprule
  \textbf{Dataset} & \textbf{Checkpoint} & \textbf{Iters} & \textbf{Search} & \textbf{Scrape} & \textbf{Python} & \textbf{Iters corr./wrong} \\
  \midrule
  \multirow{3}{*}{BrowseComp}
   & GLM-4.5-Air & 54.0 & 46.6 & 7.3 & 0.0 & 32.9 / 61.6 \\
   & Coding Agent & 70.6 & 47.6 & 22.1 & 0.9 & 42.8 / 85.1 \\
   & $+$ Search Agent & 75.3 & 51.9 & 22.2 & 1.0 & 42.5 / 92.0 \\
  \midrule
  \multirow{3}{*}{Seal-0}
   & GLM-4.5-Air & 18.0 & 11.3 & 6.4 & 0.3 & 17.9 / 18.0 \\
   & Coding Agent & 31.6 & 15.3 & 15.5 & 0.8 & 26.9 / 36.3 \\
   & $+$ Search Agent & 29.8 & 13.8 & 14.9 & 1.0 & 25.0 / 34.4 \\
  \midrule
  \multirow{3}{*}{HLE-Verified}
   & GLM-4.5-Air & 20.9 & 13.5 & 5.3 & 2.1 & 18.1 / 22.0 \\
   & Coding Agent & 33.6 & 18.1 & 12.5 & 3.0 & 30.4 / 37.0 \\
   & $+$ Search Agent & 28.8 & 14.4 & 11.0 & 3.3 & 25.6 / 32.0 \\
  \bottomrule
  \end{tabular}
  \end{table}

\paragraph{Tool-call pattern analysis.}
% Data: GLM-4.5-Air rows = message-level trajectories from the 2026-08 eval waves under the
% standardized harness. Coding-Agent RL rows = cif89awm68-seta-iter20 (the search stage's init),
% tool statistics pooled over its clean rounds (Seal-0 x5, HLE x4, BrowseComp x3), n = 6,600.
% Search-Agent RL rows = apl20-0820-iter29, tool
% statistics pooled over its clean rounds (Seal-0 x3, HLE x3, BrowseComp x2; 08-24 base + 09-02
% reruns), n = 4,575 deduped non-aborted trajectories; script eval/gl/toolstats.py, write-up
% RufusGym/docs/2026-09-18-apl20-i29-tool-usage.md (failure-mode counts in its appendix).
% GLM-4.5-Air BrowseComp rows cover 950/1,266 tasks (one shard rerunning).

Table~\ref{tab:tool_patterns} reports tool usage from every evaluation trajectory. The public GLM-4.5-Air release is search-dominated and rarely reads a page (BrowseComp search:scrape $= 6.4{:}1$): it skims snippets. By the Coding Agent checkpoint, the policy scrapes two to three times as many pages on every benchmark, shifting from skimming snippets to reading sources. Search Agent then adjusts the mix per benchmark rather than scaling everything up: on BrowseComp it adds searches at unchanged scraping, whereas on Seal-0 and HLE-Verified it searches less than its starting checkpoint and uses Python slightly more. The resulting search:scrape ratio is about 2.3:1 on BrowseComp, 1.3:1 on HLE-Verified, and 0.9:1 on Seal-0.

Depth grows along the pipeline on BrowseComp (54.0 $\to$ 70.6 $\to$ 75.3 iterations), the one benchmark whose tasks reward long multi-hop chains; on Seal-0 and HLE-Verified the Search Agent checkpoint is shallower than its init (29.8 vs.\ 31.6 and 28.8 vs.\ 33.6), so its gains there come from a different call mix at lower cost rather than from more calls. On BrowseComp, depth beyond a point stops paying: accuracy is 88\% for trajectories that finish within 25 iterations, 64\% for 25--50, and below 30\% beyond 50; wrong answers come from trajectories more than twice as deep as correct ones (92 vs.\ 43 iterations); and about 10\% of trajectories per run end by exhausting the token budget or overflowing the context window, essentially none of them correct. Past roughly 50 iterations, additional depth signals a stuck search rather than progress.

\section{Tau2-Bench with Different User Simulators}
\label{sec:tau2_usersim_appendix}

Every Tau2-Bench episode involves a simulated customer, making the user simulator part of the evaluation protocol. Table~\ref{tab:tau2_usersim} evaluates four models across three domains under three user simulators. The Rufus-Air column is an earlier checkpoint, not the final one in Table~\ref{tab:main_comparison}; the final checkpoint was evaluated under Sonnet~5 only (85.3 / 84.0 / 93.0), so this column shows how the scores move with the simulator rather than the final model's numbers. Changing the simulator shifts scores by up to $23.7$ points (GLM-4.5-Air on retail: $57.0$ under Sonnet~4 versus $80.7$ under Sonnet~5), with Sonnet~5 yielding the highest scores in all twelve model--domain pairs. These differences highlight the need to match simulator settings when comparing results across papers. Rankings are more stable: the full ordering is unchanged in telecom, and Rufus-Air ranks first in retail under all three simulators, albeit by only $0.2$ points under Sonnet~5. The airline leader varies: Nemotron-3-Super leads under Sonnet~4 and Sonnet~5, while Rufus-Air leads under Sonnet~4.5.

% ChatGPT or Claude, don't edit this table's numbers! they're settled.
\begin{table}[!ht]
\centering
\caption{Tau2-Bench pass@1 under three user simulators; Sonnet~5 is the setting used in Table~\ref{tab:main_comparison}. Only the user simulator model changes down a block; tasks, tools, the judge (Claude Sonnet~4.5), and the agent's sampling ($n{=}4$ episodes per task on the Tau2-Bench base split, temperature 1.0) are fixed. \textsuperscript{\dag}An earlier checkpoint than the final Rufus-Air in Table~\ref{tab:main_comparison}.}
\label{tab:tau2_usersim}
\setlength{\tabcolsep}{6pt}
\small
\begin{tabular}{@{}ll *{4}{c}}
\toprule
\textbf{Domain} & \textbf{User Simulator} & \textbf{\textcolor{rufusorange}{Rufus}-Air}\textsuperscript{\dag}
& \textbf{GLM-4.5-Air} & \textbf{INTELLECT-3} & \textbf{Nemotron-3-Super} \\
\midrule
% The whole Rufus-Air column is the pre-final checkpoint (the one Table 1 carried before the
% final checkpoint was selected); the final checkpoint has only been run under Sonnet 5 (Table 1).
& Sonnet~4    & 74.0 & 69.5 & 65.0 & \textbf{78.5} \\
& Sonnet~4.5  & \textbf{76.0} & 74.0 & 70.5 & 75.5 \\
\multirow{-3}{*}{Airline}
& Sonnet~5    & 80.5 & 77.0 & 76.0 & \textbf{86.0} \\
\midrule
& Sonnet~4    & \textbf{69.1} & 57.0 & 68.4 & 68.2 \\
& Sonnet~4.5  & \textbf{84.6} & 74.8 & 73.7 & 82.2 \\
\multirow{-3}{*}{Retail}
& Sonnet~5    & \textbf{86.6} & 80.7 & 75.7 & 86.4 \\
\midrule
& Sonnet~4    & \textbf{84.4} & 31.8 & 33.3 & 62.1 \\
& Sonnet~4.5  & \textbf{75.4} & 25.7 & 28.9 & 57.5 \\
\multirow{-3}{*}{Telecom}
& Sonnet~5    & \textbf{95.4} & 32.7 & 40.8 & 72.4 \\
\bottomrule
\end{tabular}
\end{table}

\end{document}